\documentclass[journal]{IEEEtran}

\usepackage{cite}
\usepackage{amsmath}
\usepackage{amsfonts}
\usepackage{array}
\usepackage{booktabs}
\usepackage{algorithm}
\usepackage[noend]{algpseudocode}
\usepackage{color}
\usepackage{url}
\usepackage[pdftex]{graphicx}
\usepackage{hyperref}
\usepackage{bbm}

\usepackage{amsmath}
\usepackage{amssymb}
\usepackage{graphicx}
\usepackage{amsmath,epsfig,amssymb,dsfont,amsthm}
\graphicspath{{figs/}}
\usepackage{caption}
\usepackage{subcaption}
\usepackage{color}
\definecolor{comment}{rgb}{0,0,1}
\usepackage{hhline}
\usepackage{subcaption}
\DeclareMathOperator*{\argmax}{arg\, max}
\DeclareMathOperator*{\argmin}{arg\, min}

\begin{document}

		\title{Graph Laplacian mixture model}
		
		\author{Hermina~Petric~Maretic
			and~Pascal~Frossard
			\thanks{Hermina Petric Maretic and Pascal Frossard are with Signal Processing Laboratory (LTS4), Ecole Polytechnique Federale de Lausanne (EPFL), Lausanne, Switzerland. E-mail: hermina.petricmaretic@epfl.ch, pascal.frossard@epfl.ch}
		}
		
		\maketitle
		\begin{abstract}
			Graph learning methods have recently been receiving increasing interest as means to infer structure in datasets. Most of the recent approaches focus on different relationships between a graph and data sample distributions, mostly in settings where all available data relate to the same graph. This is, however, not always the case, as data is often available in mixed form, yielding the need for methods that are able to cope with mixture data and learn multiple graphs. We propose a novel generative model that represents a collection of distinct data which naturally live on different graphs. We assume the mapping of data to graphs is not known and investigate the problem of jointly clustering a set of data and learning a graph for each of the clusters.
			Experiments demonstrate promising performance in data clustering and multiple graph inference, and show desirable properties in terms of interpretability and coping with high dimensionality on weather and traffic data, as well as digit classification.
		\end{abstract}
		\section{INTRODUCTION}
		Relationships between data can often be well described with a graph structure. Although many datasets, including social and traffic networks, come with a pre-existing graph that helps in interpreting them, there is still a large number of datasets (e.g., brain activity information) where a graph is not readily available. Many graph learning techniques have been proposed in the past years \cite{dong2018learning} \cite{mateos2019connecting} to help in analysing such datasets. 
		More specifically, a lot of interest in the field of graph learning is currently focused on designing graph learning methods that can take into account prior information on the graph structure \cite{egilmez2017graph}, or different relationships between data and the underlying graph  \cite{segarra2017network}. However, most of these works only consider simple data, where all datapoints follow the same model defined with only one graph. While there are still many topics of interest in those settings, we argue that natural data often comes in more complicated forms. In fact, such data opens an entire field of unsupervised learning methods.
		A natural example of such a dataset can be found in brain fMRI data, where signals usually measure the brain activity through different brain processes. Each of these processes can be explained with a different brain functional network, with regions of interest as shared network nodes. However, it is not clear which network is activated at what time, causing the need to separate signals corresponding to different networks. 
		
		In this paper, we precisely consider data that naturally forms clusters, where signals from each of the clusters live on a different graph. This allows analysis of more complex datasets where simple graph learning methods would suffer from intertwined data and thus lose the ability to capture a meaningful graph structure. 
		In particular, we study the problem of multiple graph inference from a general group of signals that are an unknown combination of data with different structures. Namely, we propose a generative model for data represented as a mixture of signals naturally living on a collection of different graphs. As it is often the case with data, the separation of these signals into clusters is assumed to be unknown. We thus propose an algorithm that will jointly cluster the signals and infer multiple graph structures, one for each of the clusters. Our method assumes a general signal model, and offers a framework for multiple graph inference that can be directly used with a number of state-of-the-art network inference algorithms, harnessing their particular benefits. 
		Numerical experiments show promising performance in terms of both data clustering and multiple graph inference, while coping well with the dimensionality of the problem. Simulations show that our method is effective in finding interesting patters in a traffic network of New York, while it demonstrates high interpretability on a weather dataset, as well as MNIST data. 
		
		As we will deal with clustering in large dimensionalities in these settings, it is worth noting that inherently high dimensional clustering problems often suffer from the curse of dimensionality \cite{bellman2015adaptive} and poor interpretability. While imposing that data lives on a graph implicitly reduces the dimensionality of the problem, graphs also offer a natural representation for connections between data. Therefore, they provide interpretability both in terms of direct inspection of graph structure, as well as the ability to further deploy various data analysis algorithms.
		
		In the literature, the graph learning problem has been first considered as sparse precision matrix inference. Data is modelled as a multivariate Gaussian distribution, whose inverse covariance matrix reveals direct pairwise connections between nodes \cite{dempster1972covariance} \cite{friedman2008sparse}. Based on these methods, several models have been proposed to infer Gaussian mixture models with sparse precision matrices \cite{danaher2014joint} \cite{lotsi2013high} \cite{hao2016simultaneous}. All these works actually focus on inferring GMRFs (Gaussian Markov Random Fields), and do not constrain values in the precision matrix in any special way. Therefore, the resulting graphs can have both positive and negative weights, as well as self-loops, which can be difficult to interpret in practice. 
		On the other hand, graph representation constrained to a valid Laplacian matrix circumvents this problem, while opening the door to numerous data analysis methods \cite{shuman2013emerging}. For these reasons, an increasing amount of work has recently been focusing on inferring (generalised) graph Laplacians. Among the first researchers to focus on graph Laplacian inference, Dong et al. \cite{Dong14} adopt a graph signal processing perspective to enforce data smoothness on the inferred graph. The proposed model results in assumptions similar to those in GMRFs, but with added constraints that ensure that the graph is given by a valid Laplacian matrix. Kalofolias \cite{kalofolias2016learn} uses a similar framework and proposes a computationally more efficient solution by inferring a weight matrix, which can eventually be easily transformed into a Laplacian. An even more efficient, approximate solution has been proposed for large scale graph learning \cite{kalofolias2017large}, scaling to graphs with millions of nodes. Other recent works in this vein include inference of graph shift operators, with the assumption that data is the result of a graph diffusion process. A popular approach to this problem consists in exploiting the fact that the eigenvectors of the graph will be shared with those of any graph filter. Therefore, they can be estimated from the data sample covariance matrix, and the optimisation can be done only over the eigenvalues \cite{segarra2017network} \cite{pasdeloup2017characterization}. 
		Dictionary based methods try to model signal heterogeneity by taking into account sparse combinations or dictionary atoms. In \cite{thanou2017learning}, signals are represented as linear combinations of atoms from a heat kernel dictionary. As those are still bound to be smooth, \cite{maretic2017graph} model signals as sparse linear combinations of atoms in a predefined polynomial graph dictionary. 
		Several methods exploit possible additional priors in the graph inference problem, such as a bandlimited representation of signals \cite{sardellitti2019graph}, or properties on graph topologies \cite{lu2018learning} \cite{pavez2018learning}.
		  All aforementioned methods assume the whole set of signals can be well explained with one single graph model. Naturally, this is unlikely to be the case in many applications, where signals might be generated in different ways or exist as a combination of distinct causes.
		
		Finally, there have been some works focused on signals in one dataset that naturally live on different graphs. Kalofolias et al. \cite{kalofolias2017learning} infer the structure of a time varying graph, effectively capturing multiple graphs in different periods of time. A similar approach based on sparseness of variation between graphs in different periods has been proposed by Yamada et al. \cite{yamada2019time}. Sardellitti et al. \cite{sardellitti2019enabling} propose a method for multi-layer graph inference for prediction of dynamic graph signals. Segarra et al. \cite{segarra2017joint} infer multiple networks under the assumption of signal stationarity. Works inspired by graphical lasso \cite{friedman2008sparse} include a method by Kao et al. \cite{kao2017disc} that promotes differences in inferred graphs, minimizing the correlation between each graph and the sample covariance of signals that do not live on it. Recent work of Gan et al. \cite{gan2019bayesian} imposes that the sparsity pattern on all inferred graphs should be similar through a Bayesian prior. However, differently from our work, all of these methods assume that signal clusters are given a priori, i.e., it is clear which signals correspond to which graph. The joint network inference then becomes a problem of imposing similarities on different graph structures, rather than decoupling the signals into groups and learning a graph to explain each of the groups, which is the focus of our method. To the best of our knowledge, this work presents the first framework to deal with inference of multiple graph Laplacians from mixed (not a priori clustered) signals, part of which (the heat kernel model) has been published at \cite{maretic2018graph}.

		\section{GRAPH SIGNAL PROCESSING BASICS} \label{basics}
		Let $G = (V, E, \boldsymbol W)$ be an undirected, weighted graphs with a set of $N$ vertices $V$, edges $E$ and a weighted adjacency matrix $\boldsymbol W$. The value $W_{ij}$ is equal to $0$ if there is no edge between $i$ and $j$, and denotes the weight of that edge otherwise. The non-normalised (or combinatorial) graph Laplacian is defined as 
		\begin{align} \label{eq:lapl}
		\boldsymbol L = \boldsymbol D - \boldsymbol W,
		\end{align} 
		where $\boldsymbol D$ is a diagonal matrix of node degrees. A graph Laplacian satisfies:
		\begin{align}
		L_{i,j} = L_{j,i} &\leq 0, \forall i \neq j,\\
		\sum_{j=1}^{N} L_{i,j} &= 0,  \forall i.
		\end{align}
		When these conditions are satisfied, we write $\boldsymbol L\in \mathcal{L}$, where $\mathcal{L}$ is a set of valid Laplacian matrices \cite{shuman2013emerging}. 
		
		We also define a signal on a graph as a function $x:V \rightarrow \mathbb{R}$, where $x_v$ denotes the value of a signal on a vertex $v$. A graph signal is considered smooth if most of its energy is concentrated in the low frequencies of the underlying graph, which can be measured with a quadratic form of the graph Laplacian:
		\begin{align} \label{eq:smoothness}
		\boldsymbol x^T\boldsymbol L \boldsymbol x = \frac{1}{2} \sum_{i,j} W_{ij} (x_i - x_j)^2.
		\end{align}
		Indeed, it is clear from Equation (\ref{eq:smoothness}) that the signal difference will get more penalised for two vertices linked by a strong edge. It might be less apparent that there is also a strong spectral interpretation. Namely, as a real symmetric matrix, $\boldsymbol L$ can be decomposed into a set of $N$ orthogonal eigenvectors and associated eigenvalues. These play the role of Fourier transform in graph signal processing, with the eigenvalues taking up a notion associated to frequency. Using this decomposition with $\boldsymbol U$ being the eigenvectors and $\boldsymbol \Lambda$ diagonal matrix of eigenvalues, we can see that the above relation
		\begin{align}
		\boldsymbol x^T\boldsymbol L \boldsymbol x= \boldsymbol x^T\boldsymbol U \boldsymbol\Lambda \boldsymbol U^T \boldsymbol x
		\end{align}
		penalises signals according to their frequency support. 
		
		Given a graph filter $g(\boldsymbol L)$ \cite{ortega2018graph} and a white noise signal $\boldsymbol w \sim \mathcal{N}(0, \mathbb{I})$, we define a kernel graph signal as $\boldsymbol x = \boldsymbol \mu + g(\boldsymbol L) \boldsymbol w$. The kernel signal will follow a Gaussian distribution:
		\begin{align}
			\boldsymbol x =\boldsymbol \mu + g(\boldsymbol L)\boldsymbol w \sim \mathcal{N}(\boldsymbol \mu, g(\boldsymbol L) \mathbb{I} g(\boldsymbol L)^T) = \mathcal{N}(\boldsymbol \mu, g^2(\boldsymbol L))
			\label{eq:kernelModel}
		\end{align}
		Notice that a smooth signal can be seen as a special case of the kernel signal, with the filter equal $g(\boldsymbol L) = \sqrt{\boldsymbol L^\dagger}$, $\boldsymbol L^\dagger$ being the pseudo-inverse of the graph Laplacian matrix. Namely, assuming the signal can be modelled through a latent variable $\boldsymbol h \sim \mathcal{N}(0,\boldsymbol \Lambda ^\dagger)$, such that $\boldsymbol x = \boldsymbol \mu +\boldsymbol U \boldsymbol h$, yields the minimiser of \ref{eq:smoothness} w.r.t $\boldsymbol L$ as a maximum likelihood estimator for the graph $\boldsymbol L$. As $\boldsymbol \Lambda ^\dagger$ represents the pseudo-inverse of the eigenvalue matrix, this model assumes that signal support is inversely proportional to frequency, describing smooth signals. This gives us a direct relationship between signals and the graph Laplacian matrix $\boldsymbol L$:
		\begin{align}
		\boldsymbol x = \boldsymbol \mu + \boldsymbol U\boldsymbol h \sim \mathcal{N}(\boldsymbol \mu, \boldsymbol U\boldsymbol \Lambda^\dagger \boldsymbol U^T) = \mathcal{N}(\boldsymbol \mu, \boldsymbol L^\dagger). \label{eq:smoothGauss}
		\end{align}
		Note that $x$ can also be seen as a special degenerate GMRF, since $L_{ij} = 0 \Leftrightarrow (i,j)\notin E$. Namely, the precision matrix $\boldsymbol L$ has at least one zero eigenvalue, as well as a special structure ensuring $\boldsymbol L \in \mathcal{L}$. 
		
		 As this special case is of large importance and brings specific challenges, we will explore both the general, and this special case, in detail. 
		
		\section{GRAPH LAPLACIAN MIXTURE MODEL}
\begin{figure}[h]
			\centering
			\includegraphics[scale = 0.2]{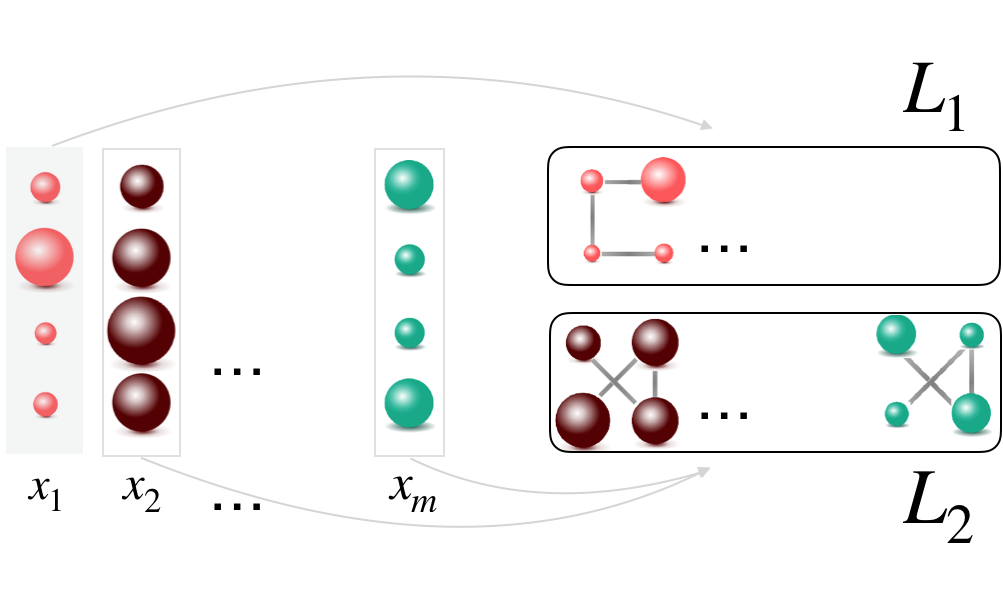}
			\caption{Illustration of our model. Data is given as signals $\boldsymbol x_1,...,\boldsymbol x_m$. In this example, each signal is smooth on its respective graph. Our model groups signals naturally belonging to different clusters and learns a graph for each of the clusters.}\label{fig:glmm_toy}
		\end{figure}

		We propose a probabilistic model for a set of graph signals with distinguishable subsets, where the behaviour of signals in each of the subsets (groups) is well explained with a graph. 
		A toy example of our data model is given in Figure \ref{fig:glmm_toy}. 
	
		Our goal is to distinguish these groups of signals, and eventually infer a graph that will model the structure of signals in each cluster. The clusters of signals are unknown and the graph structures largely influence behaviours inside clusters. We thus argue that identifying the clusters and learning the associated graphs are intertwined problems. Therefore, we propose a generative model that jointly explains both signals and clusters, under an assumption of signals following the graph kernel model, for each cluster. 
		
		\subsection{Multigraph signal representation}
		
		\begin{figure}[h]
			\centering
			\includegraphics[scale = 0.3]{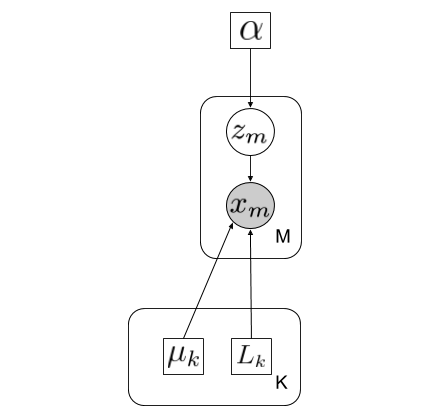}
			\caption{Plate notation for our generative model. Filled in circles are observed variables, small empty squares are unknown parameters, and the empty circles represent latent variables. Large plates indicate repeated variables.}\label{fig:glmm_plate}
		\end{figure}
		
		The graphical representation for our model is given in Figure \ref{fig:glmm_plate}. Let us assume that there are $K$ undirected, weighted graphs $G^k = (V, E^k, \boldsymbol W^k)$ with a set of $N$ shared vertices $V$. Each graph has a specific set of edges $E^k$ and a weighted adjacency matrix $\boldsymbol W^k$. From each of these weight matrices $\boldsymbol W^k$, we can define a graph Laplacian matrix $\boldsymbol L_k$, as in  (\ref{eq:lapl}). 
		
		We further assume there are $K$ clusters, and each of the $M$ observed signals $\boldsymbol x_m \in \mathbb{R}^N$ on the nodes $V$, belongs to exactly one of the clusters. Cluster participation is modelled through a binary latent variable $ \boldsymbol z_{m} \in \mathbb{R}^K$, with $z_{m, j} = \delta_{j,k}, \forall j$, if signal $\boldsymbol x_m$ belongs to cluster $k$. Mixing coefficients $\boldsymbol \alpha \in \mathbb{R}^K$ model the size of each cluster, and define a prior distribution on variables $\boldsymbol z_m$, with $p(z_{m,j} = \delta_{j,k}, \forall j) = p(z_{m,k} = 1)= \alpha_k, \forall m$. 
		
		Finally, we model data in each cluster $k$ with a mean $\boldsymbol \mu_k$ and a graph Laplacian $\boldsymbol L_k$, assuming associated signals will be close to $\boldsymbol \mu_k$ and follow a kernel model with a filter $g$ on graph $\boldsymbol L_k$, as defined in Eq. (\ref{eq:kernelModel}):
		\begin{align}
		p(\boldsymbol x_m|z_{m,k} = 1) = p(\boldsymbol x_m|\boldsymbol \mu_k, g_k(\boldsymbol L_k)) = \mathcal{N}(\boldsymbol \mu_k,g_k^2(\boldsymbol L_k))
		\end{align}
		Marginalising over latent variables $\boldsymbol{z}$, we have:
		\begin{align}
		p(\boldsymbol x_m) = &\sum_{\boldsymbol z_m} p(\boldsymbol z_m) p(\boldsymbol x_m|\boldsymbol z_m)\\
		=&\sum_{k=1}^K p(z_{m,k} = 1) p(\boldsymbol x_m|z_{m,k} = 1)\\
		=  &\sum_{k=1}^K \alpha_k \mathcal{N}(\boldsymbol \mu_k, g_k^2(\boldsymbol L_k)), \label{eq:model_l1}\\
		\text{s.t. } \quad
		&\boldsymbol L_k \in \mathcal{L}, \forall k \label{eq:model_l2}\\
		&\sum_{k=1}^K \alpha_k = 1, \label{eq:model_l3} \\
		& \alpha_k > 0, \forall k \label{eq:model_l4} 
		\end{align} 
This fully describes our generative model, with (\ref{eq:model_l2}) ensuring that all $\boldsymbol L_k$s are valid Laplacians, while (\ref{eq:model_l3}) and (\ref{eq:model_l4}) ensure that $\boldsymbol \alpha$ defines a valid probability measure.
		\subsection{Problem formulation}
		Given a set of $M$ $N$-dimensional graph signals $\boldsymbol X \in \mathbb{R}^{N \times M}$ with some intrinsic grouping into $K$ clusters associated to it, we look at the MAP estimate for our parameters: mixing coefficients $\boldsymbol \alpha = \alpha_1...\alpha_K$, means $\boldsymbol\mu = \boldsymbol \mu_1...\boldsymbol \mu_K$ and graph Laplacians $\boldsymbol L = \boldsymbol L_1...\boldsymbol L_K$. Namely, assuming the data has been sampled independently from the distribution in (\ref{eq:model_l1}), and allowing for a prior on the graph structure, we want to maximise over the a posteriori distribution of our model:
		\begin{align}
		&\argmax_{\boldsymbol \alpha, \boldsymbol\mu, \boldsymbol L} \enskip \text{ln } p(\mathbf{\boldsymbol\alpha}, \boldsymbol\mu, \boldsymbol L | \boldsymbol X) \\
		\propto &\argmax_{\boldsymbol \alpha, \boldsymbol\mu, \boldsymbol L} \enskip \text{ln } p(\boldsymbol X|\mathbf{\boldsymbol\alpha}, \boldsymbol\mu, \boldsymbol L) p(\boldsymbol L)\\
		= &\argmax_{\boldsymbol \alpha, \boldsymbol\mu, \boldsymbol L} \enskip \text{ln } \prod_{m=1}^M p(\boldsymbol x_m|\mathbf{\boldsymbol\alpha}, \boldsymbol\mu, \boldsymbol L) p(\boldsymbol L)\\
		= &\argmax_{\boldsymbol \alpha, \boldsymbol\mu,\boldsymbol L} \enskip  \text{ln } \prod_{m=1}^{M} \sum_{k=1}^K \alpha_k \mathcal{N}(\boldsymbol x_m|\boldsymbol \mu_k, g_k^2(\boldsymbol L_k))  p(\boldsymbol L_k) \label{eq:ll_1}\\
		=&\argmax_{\boldsymbol\alpha, \boldsymbol\mu, \boldsymbol L} \enskip  \sum_{m=1}^{M} \text{ln }  \sum_{k=1}^K \alpha_k \mathcal{N}(\boldsymbol x_m|\boldsymbol \mu_k, g_k^2(\boldsymbol L_k))p(\boldsymbol L_k),\label{eq:ll_2}
		\end{align}
		which does not have a closed form solution. We will thus estimate the parameters using an expectation maximisation (EM), as explained below.
		
		Note that we present our algorithm for the case of general kernel signals as long as that is possible. However, we will see that smooth signals pose specific challenges, and will for that reason be treated separately throughout the paper.
		\section{ALGORITHM}
		We propose an expectation maximisation algorithm that alternates between optimising for expected cluster participations $\boldsymbol\gamma: =\mathbb{E} (\boldsymbol z)$ in the expectation step, and signal means $\boldsymbol\mu$, class proportions $\boldsymbol\alpha$ and graph topologies $\boldsymbol L$ in the maximisation step. Therefore, the joint clustering and multi-graph learning problem iterates over two steps: the first one estimates the correct clustering, and the second one describes the clusters by inferring cluster means and proportions, as well as the graphs describing them. Precisely, we first initialise $\boldsymbol\alpha, \boldsymbol\mu$ and $\boldsymbol L$ randomly, noting that we ensure $\boldsymbol L$ are set to a random valid Laplacian matrix, guaranteeing it truly describes a graph structure. The alternating steps follow, as described below:
		\subsection{Expectation (E step)}
		Let us define $\boldsymbol \gamma \in \mathbb{R}^{M \times K}$ as a matrix of posterior probabilities, with $\gamma_{m,k}$ modelling the probability that the signal $\boldsymbol x_m$ belongs to the group $k$. Note that, at the same time, this is the expected value of the latent indicator variable $z_{m,k}$, with $\boldsymbol \gamma_m  =\mathbb{E} (\boldsymbol z_m)$.
		\begin{align} 
		\gamma_{m,k} &= p(z_{m,k} = 1 |\boldsymbol x_m,\boldsymbol  \mu_k, \boldsymbol L_k) \\
		&= \frac{p(z_{m,k} = 1) p(\boldsymbol x_m| z_{m,k} = 1, \boldsymbol \mu_k, \boldsymbol L_k)}{\sum_{l=1}^K p(z_{m,l} = 1) p(\boldsymbol x_m| z_{m,l} = 1, \boldsymbol \mu_l, \boldsymbol L_l)}\\
		&= \frac{\alpha_k \mathcal{N}(\boldsymbol x_m|\boldsymbol \mu_k, g_k^2(\boldsymbol L_k))}{\sum_{l=1}^K \alpha_l \mathcal{N}(\boldsymbol x_m|\boldsymbol \mu_l,g_l^2(\boldsymbol L_l))} \label{eq:gamma}
		\end{align}
		
		In practice, $\boldsymbol \gamma$ can take different forms, depending on the choice of kernels $g_k, k\in \{1,..,K\}$. In particular, two common models are the heat kernel model and the general smooth graph signal model.
		
		\subsubsection{The heat kernel signal model}
		The posterior probabilities $\boldsymbol \gamma$ can be directly computed in the case of most arbitrarily chosen graph kernels using Equation (\ref{eq:gamma}). One of those kernels is the graph heat kernel, which we will use as an example and explore in more detail throughout this paper.
		
		\subsubsection{The smooth signal model}
		This formulation for $\boldsymbol \gamma$ cannot be computed directly when signals follow the smooth model. Namely, it is well known that a graph Laplacian has at least one eigenvalue that is zero, corresponding to the eigenvector $\mathbf{1}$. This makes the distribution in (\ref{eq:gamma}) degenerate. 
		At the same time, from the signal processing point of view, the corresponding eigenvector is completely smooth on all possible graphs, and is therefore non-informative. The disintegration theorem \cite{kallenberg2006foundations}  guarantees that we can restrict our problem to the $N-1$ dimensional subspace spanned by all remaining eigenvectors. We thus proceed by projecting signals to this subspace, where we can then compute $\boldsymbol \gamma$ and retrieve the probabilities we wanted to model.
		
		Furthermore, if a graph has disconnected components, it will have as many zero eigenvalues as the number of components in the graph. Even if the final graph is connected, there is no guarantee that the algorithm does not return a disconnected graph in one of the iterations of the optimisation problem. 
		It is easy to see how this can pose large numerical problems, both in each graph separately as their eigenvalues approach zero, but also in terms of comparing the probabilities that these graphs define when trying to infer $\gamma_{m,k}$ from (\ref{eq:gamma}). To avoid this problem, we add a small regularising constant $\epsilon$ to every eigenvalue corresponding to the $N-1$ non-trivial eigenvectors of the graph Laplacian. Note that $\epsilon$ serves only for numerical regularization and should be kept as small as possible.
		Finally, with $\boldsymbol y_{m,k} :=  \boldsymbol x_m - \boldsymbol \mu_k$, the computation of probabilities $\boldsymbol \gamma$ sums up as follows:
		\begin{align}
		\boldsymbol L_k &= \boldsymbol U_k \boldsymbol \Lambda_k \boldsymbol U_k^T \\
		\tilde{\boldsymbol \Lambda}_k &= (\boldsymbol \Lambda_k)_{2:N,2:N} + \epsilon \mathbb{I} \\
		\tilde{\boldsymbol U}_k &= (\boldsymbol U_k)_{1:N,2:N} \\
		\tilde{\boldsymbol y}_{m,k} &= \tilde{\boldsymbol U}_k^T \boldsymbol y_{m,k}\\
		\gamma_{m,k} &= \frac{\alpha_k \mathcal{N}(\tilde{\boldsymbol y}_{m,k}|0, \tilde{\boldsymbol \Lambda}_k^{-1})}{\sum_{l=1}^K \alpha_l \mathcal{N}(\tilde{\boldsymbol y}_{m,l}|0, \tilde{\boldsymbol \Lambda}_l^{-1})}
		\end{align}
		\subsection{Maximisation (M step)}
		
		Having estimated probabilities $\gamma_{m,k}$ in the E-step, we can now maximise the expected posterior distribution given all observed signals, to infer $\boldsymbol \alpha, \boldsymbol\mu$ and $\boldsymbol L$:
		\begin{align}
		\argmax_{\boldsymbol \alpha, \boldsymbol\mu, \boldsymbol L} \enskip &\sum_{\boldsymbol Z} p(\boldsymbol Z | \boldsymbol X, \boldsymbol \mu, \boldsymbol L) \text{ln } p(\boldsymbol X, \boldsymbol Z|\mathbf{\boldsymbol\alpha}, \boldsymbol\mu, \boldsymbol L) p(\boldsymbol L)\\
		= \argmax_{\boldsymbol \alpha, \boldsymbol\mu, \boldsymbol L} \enskip &\sum_{\boldsymbol Z} p(\boldsymbol Z | \boldsymbol X, \boldsymbol \mu, \boldsymbol L) \\
		&\text{ln } \prod_{m=1}^{M}  p(\boldsymbol x_m, \boldsymbol z_m|\mathbf{\boldsymbol\alpha}, \boldsymbol\mu, \boldsymbol L) p(\boldsymbol L)\\
		= \argmax_{\boldsymbol \alpha, \boldsymbol\mu, \boldsymbol L} \enskip &\sum_{m=1}^{M} \sum_{k=1}^K p(z_{m,k} = 1| \boldsymbol x_m, \boldsymbol \mu_k, \boldsymbol L_k) \\
		&\text{ln }   p(\boldsymbol x_m, \boldsymbol z_m|\alpha_k, \boldsymbol \mu_k, \boldsymbol L_k) p( \boldsymbol L_k)\\
		= \argmax_{\boldsymbol \alpha, \boldsymbol\mu,\boldsymbol L} \enskip   &\sum_{m=1}^{M} \sum_{k=1}^K \boldsymbol \gamma_{m,k} \text{ln } (\alpha_k \mathcal{N}(\boldsymbol x_m|\boldsymbol \mu_k,  g_k^2(\boldsymbol L_k))  p(\boldsymbol L_k)) \\
		= \argmax_{\boldsymbol \alpha, \boldsymbol\mu,\boldsymbol L} \enskip   &\sum_{m=1}^{M} \sum_{k=1}^K \gamma_{m,k} (\text{ln } \alpha_k +\\
		&+ \text{ln } \mathcal{N}(\boldsymbol x_m|\boldsymbol \mu_k,  g_k^2(\boldsymbol L_k)) +  \text{ln } p(\boldsymbol L_k)) \label{eq:map_2}
		\end{align}
		
		Equation (\ref{eq:map_2}) is concave over $\boldsymbol \mu$ and $\boldsymbol  \alpha$, and offers closed form solutions for both:
		\begin{align} 
		\boldsymbol \mu_k &= \frac{\sum_{m=1}^{M}\gamma_{m,k} \boldsymbol x_m}{\sum_{m=1}^{M} \gamma_{m,k} } \label{eq:mu}\\
		\alpha_k &= \frac{\sum_{m=1}^{M} \gamma_{m,k}}{N }. \label{eq:alpha}
		\end{align}
		In order to maximise over $\boldsymbol L$, we substitute $\boldsymbol x_m$ with variables $\boldsymbol y_{m,k} :=  \boldsymbol x_m - \boldsymbol \mu_k$. Now we can formulate a problem of multiple graph inference:
		\begin{align}
		\argmax_{\boldsymbol L} \enskip   \sum_{k=1}^K \sum_{m=1}^{M} \gamma_{m,k} (\text{ln } \mathcal{N}(\boldsymbol y_{m,k}|\boldmath 0, g_k^2(\boldsymbol L_k)) +  \text{ln } p(\boldsymbol L_k)).
		\end{align}
		It is clear that these are $K$ independent optimisation problems, and we can maximise each one separately:
		\begin{align}
		\argmax_{\boldsymbol L_k \in \mathcal{L}} \enskip   \sum_{m=1}^{M} \gamma_{m,k} (\text{ln } \mathcal{N}(\boldsymbol y_{m,k}|\boldmath 0, g_k^2(\boldsymbol L_k)) +  \text{ln } p(\boldsymbol L_k)). \label{eq:gl_general_kernel}
		\end{align}
		The generative model gives a general framework and a graph inference algorithm of choice can be used to infer $\boldsymbol{L}$, depending on the nature of data and the appropriate graph filter. Here, we explore two different methods, one representing a special case of smooth graph signals, while the other investigates one of the most used graph filters.
		\subsubsection{Graph inference from smooth signals}
		Similarly to the graph inference problem explored in \cite{Dong14}, using a kernel $g(\boldsymbol L_k) = \sqrt{\boldsymbol L^\dagger}$, each of these problems can be efficiently approximated with:
		\begin{align}
		\argmin_{\boldsymbol L_k \in \mathcal{L}} \enskip   \sum_{m=1}^{M} \gamma_{m,k} \boldsymbol y_{m,k}^T \boldsymbol L_k \boldsymbol y_{m,k} +  f_k(\boldsymbol L_k), \label{eq:gl_general}
		\end{align}
		with the graph prior encoded in $f_k(\boldsymbol L_k)$.
		
		To efficiently update $\boldsymbol L$ through (\ref{eq:gl_general}), it is crucial to choose a good prior on the graph Laplacians. Namely, we want to maximise signal smoothness on the graph, while controlling graph sparsity. Due to its computational efficiency, we will use the algorithm from Kalofolias \cite{kalofolias2016learn} with the small addition of cluster probabilities $\gamma_{m,k}$, as in (\ref{eq:gl_general}). The graph update step thus becomes:
		\begin{align} \label{eq:gl}
		\argmin_{\boldsymbol L_k \in \mathcal{L}} \quad & \sum_{m=1}^{M} \gamma_{m,k} \boldsymbol y_{m,k}^T \boldsymbol L_k \boldsymbol y_{m,k} - \\
		& \beta_{1,k} tr(\boldsymbol{1}^T log (\text{diag}(\boldsymbol L_k))) + \beta_{2,k} \| \boldsymbol L_k \|_{F,\text{off}}^2 \\
		\mathcal{L} = \{\tilde{\boldsymbol L} \in &\mathbb{R}^{N\times N} : \tilde{L}_{i,j} = \tilde{L}_{j,i} \leq 0, \forall i \neq j, \tilde{\boldsymbol L} \boldsymbol 1 = \boldsymbol 0\} 
		\end{align}
		in which diag($\boldsymbol L_k$) is a vector with the diagonal values (node degrees) from $\boldsymbol L_k$, and $\| \boldsymbol L_k \|_{F,\text{off}}^2$ is the Frobenius norm of the off-diagonal values in $\boldsymbol L_k$. Notice that $\| \boldsymbol L_k \|_{F,\text{off}}^2$ = $\| \boldsymbol W_k \|_{F}^2$, where $\boldsymbol W_k$ is the weight matrix of the same graph. Compared to the formulation from (\ref{eq:gl_general}), the function $f_k(\boldsymbol L_k) = - \beta_{1,k} tr(\boldsymbol{1}^T log (\text{diag}(\boldsymbol L_k))) + \beta_{2,k} \| \boldsymbol L_k \|_{F,\text{off}}^2 $ consists of two parts, such that increasing $\beta_{1,k}$ strengthens graph connectivity, while decreasing $\beta_{2,k}$ promotes sparsity. They can be selected with a simple parameter sweep, or automatically as proposed in \cite{kalofolias2017large}. Note that this $f_k(\boldsymbol L_k)$ does not stem from a probabilistic prior, and is used here as an effective heuristic to approximate problem (\ref{eq:gl_general}). As before, the constraint that $\boldsymbol L_k \in \mathcal{L}$ ensures that $\boldsymbol L_k$ is a valid Laplacian matrix. This problem is solved with an iterative algorithm \cite{kalofolias2016learn} and the computational complexity of this algorithm is $\mathcal{O}(MN^2) $ once for preprocessing, and then $\mathcal{O}(N^2) $ per iteration. 
		
		\subsubsection{Graph inference from kernel signals}
		Assuming the filter is known a priori, an efficient graph inference method can easily be formulated. As an example, we investigate a widely used filter, the graph heat kernel,
		\begin{align}
		g_k(\boldsymbol L_k) = e^{-\tau  \boldsymbol L_k}.
		\end{align}
		
		Following (\ref{eq:gl_general_kernel}), it is clear that data is connected to the graph Laplacian through its covariance $g_k^2(\boldsymbol L_k) = e^{-2\tau \boldsymbol L_k}.$ To infer a graph Laplacian matrices $\boldsymbol L_k$, we first notice that the estimation of a covariance matrix without graph priors $p(\boldsymbol L_k)$ can be written in closed form as:
		\begin{align}
		\boldsymbol \Sigma_k : = \frac{\sum_m \gamma_{m,k}\boldsymbol y_{m,k}\boldsymbol y_{m,k}^T}{\sum_m \gamma_{m,k}}
		\end{align}
		In order to efficiently infer graph structures, the information of data probability might however not be sufficient. Namely, without very large amounts of data, the sample covariance matrices $\{\boldsymbol \Sigma_k\}$ are usually noisy (if not low rank), and it can be difficult to recover the exact structure of the graph matrix. We thus formulate a problem that aims at finding a valid Laplacian matrix that would give a covariance matrix similar to the sample covariance one, while at the same time imposing a graph sparsity constraint. Namely, we can estimate the weight matrix $\boldsymbol W_k$ as
		\begin{align}
		& \qquad \argmin_{W_k \in \mathcal{W}} \|\boldsymbol \Sigma_k - e^{-2\tau \boldsymbol L_k}\|_F^2 + \beta_k \|\boldsymbol W_k\|_1,  \\
		\rm{s.t.} & \left\{\begin{aligned}  & \boldsymbol L_k = \boldsymbol D_k - \boldsymbol W_k \\ & 
		\mathcal{W} = \{\boldsymbol W_k \in \mathbb{R}_+^{N\times N} : \boldsymbol W_k = \boldsymbol W_k^T, diag(\boldsymbol W_k) = 0\} 
		\end{aligned}\right.\nonumber
		\end{align}
		This is equivalent to solving 
		\begin{align}
		& \argmin_{\boldsymbol W_k \in \mathcal{W}} \|\text{log} \boldsymbol \Sigma_k + 2\tau \boldsymbol L_k\|_F^2 + \beta_k \|\boldsymbol W_k\|_1 
		\end{align}
		with the same constraints. It results in a convex problem that can be solved efficiently with FISTA \cite{beck2009fast} \cite{maretic2018graph}.

		As our work offers a generic model for multiple graph inference, a very wide range of graph signal models, and most graph learning methods can be used directly with our framework to perform multiple graph inference.
		Moreover, while performing graph inference already has a clear advantage with respect to covariance estimation in terms of both interpretability and accuracy \cite{dempster1972covariance}, this effect can be enhanced using graph-related prior information to simplify the problem. Namely, any graph specific additional information, such as a mask of possible or forbidden edges, or a desired level of sparsity, can easily be incorporated in most graph learning methods, rendering the implicit dimensionality reduction of the problem even stronger, and the search space smaller, thus leading to even better results.
		\section{SIMULATIONS}
		In this section, we evaluate our algorithm on both synthetic and real data. To the best of our knowledge, there are no other methods tackling the problem of jointly clustering graph signals and learning multiple graph Laplacians. Thus, we compare our algorithm (namely, GLMM for the smooth signal model, GHMM for the heat kernel) to the estimation of a Gaussian mixture model (GMM) and respectively to a signal clustering by a simple K-means algorithm, followed by graph learning \cite{kalofolias2016learn} in each inferred cluster separately (denoted as K-means + GL). In order to compare the inferred graphs, we threshold all graphs obtained with a graph learning method to remove very small values. Furthermore, as the Gaussian mixture model does not contain any sparsity regularization, and will thus not naturally provide almost sparse inverse Gaussians, we obtain graphs by choosing only the largest absolute values in the inferred precision matrices, such that the number of edges is the same as the one in the original graph.
		\subsection{Synthetic data}
		\subsubsection{Smooth signals on a graph mixture}
		We consider randomly generated connected graphs of size $N=15$ following an Erdos-Renyi model \cite{erdds1959random} with $p=0.7$. Each graph $\boldsymbol L_k$ is given a randomly generated 15-dimensional mean signal $\boldsymbol \mu_k$ that lives on its vertices, $\boldsymbol \mu_k \sim \mathcal{N}(0, \sigma^2 \mathbb{I})$, with $\sigma = 0.5$. We fix the total number of signals to 150 and consider a case with 2 and 3 classes, given by the probability vectors $\boldsymbol  \alpha^1 = \{0.5, 0.5\}$, $\boldsymbol  \alpha^2 = \{0.2, 0.8\}$ and $\boldsymbol  \alpha^3 = \{0.33, 0.33, 0.33\}$. For each signal $\boldsymbol x_m$ we randomly generate $\boldsymbol z_m$ through probabilities $\boldsymbol  \alpha$. Then $\boldsymbol x_m$ is randomly generated through a distribution $\boldsymbol x_m \sim \mathcal{N}(\boldsymbol \mu_k, \boldsymbol L_k^\dagger)$, if $z_{m,k} = 1$, which gives us the full synthetic dataset for each experiment. The whole experiment has been repeated 100 times, each time with different graphs, means and randomly generated data.
		
		We examine the performance of GLMM, GMM and K-means + GL in recovering the original clusters of signals (given by $\boldsymbol z_m$) and the corresponding graph topologies. The hyperparameters have been fixed with a grid search before running the experiment.
 Note that, to avoid numerical issues and render the comparison more fair, we train the GMM in $n$ - 1 dimensions, ignoring the direction of the first eigenvector, as the corresponding eigenvalue is known to be zero (see Section \ref{basics}). We present the error in terms of class identification NMSE ($\frac{1}{2M} \| \boldsymbol z - \boldsymbol \gamma\|_F^2$, presented in $\%$) in Table \ref{tb:syn}. The performance of graph inference for each of the methods is presented in Table \ref{tb:syn_gr} in terms of mean edge recovery F measure.
		
		\begin{table}[h]
			\caption{Synthetic data clustering results for graphs with 15 nodes. The error is presented in terms of signal clustering NMSE (\%). The first row shows results for 2 balanced clusters with $\boldsymbol \alpha = [0.5, 0.5]$, the second for 3 balanced clusters $\boldsymbol \alpha = [0.33, 0.33, 0.33]$, while the last row presents the error for clustering unbalanced clusters with $\boldsymbol \alpha = [0.2, 0.8]$.}
			\centering
			\begin{tabular}{|c||c|c|c|c|}
				\hline
				$\boldsymbol \alpha$ &	GLMM & GMM  & K-Means \\ 
				\hhline{|=||=|=|=|}
				[0.5, 0.5]  & \textbf{2.49}  & 22.6 & 7.3  \\
				\hline
				[0.33, 0.33, 0.33]  & \textbf{5.98}  & 27.7 & 11.86  \\
				\hline
				[0.2, 0.8]   & \textbf{2.84} & 23.02 & 21.03 \\ \hline
			\end{tabular}\label{tb:syn}
		\end{table}
		
		\begin{table}[h]
			\caption{Synthetic data graph learning results for graphs with 15 nodes. The performance is evaluated in terms of edge recovery F-measure. The first row shows results for balanced clusters with $\boldsymbol \alpha = [0.5, 0.5]$, the second for 3 balanced clusters $\boldsymbol \alpha = [0.33, 0.33, 0.33]$, while the last two rows present the F-measure for unbalanced clusters with $\alpha_1 = 0.2$, and $\alpha_2 = 0.8$, respectively.}
			\centering
			\begin{tabular}{|c||c|c|c|}
				\hline
				$\boldsymbol \alpha$ &	GLMM & GMM  & K-Means + GL \\ 
				\hhline{|=||=|=|=|}
				[0.5, 0.5]                               & \textbf{0.81 }                                   & 0.59 & 0.77  \\ \hline
				[0.33, 0.33, 0.33]                               & \textbf{0.71 }                                   & 0.44 & 0.64  \\ \hline
				0.2                                      & \textbf{0.66}                                     & 0.39 & 0.52  \\ \hline
				0.8                                    & \textbf{0.86}                                     & 0.7 & 0.81  \\ \hline
			\end{tabular}\label{tb:syn_gr}
		\end{table}
		
		As expected, we can see that all methods are affected by unbalanced clusters, and give significantly poorer results in terms of clustering performance, as well as edge recovery for the graph in less represented cluster. As the graph in the more represented cluster (with $\alpha_2 = 0.8$) will always be inferred from the bigger set of relevant signals, all three methods outperform their own F measure score compared to the case of balanced clusters. 
		
		\subsubsection{Signals on a mixture of graph heat kernels}
		We next investigate the performance of our algorithm for signals generated from a graph filter. 
		
		We generate two connected Erdos-Renyi graphs $\boldsymbol L_1$ and $\boldsymbol L_2$ of size $N = 20$, with edge probability $p = 0.7$. The means for each cluster are randomly drawn from $\boldsymbol \mu_k \sim \mathcal{N}(0,0.1 \mathbb{I})$, and the membership probabilities for each cluster are fixed to $\alpha_k = 0.5$. The signals are random instances of Gaussian distributions $\boldsymbol x_m \sim \mathcal{N}(\boldsymbol \mu_k, e^{-2\tau \boldsymbol L_k})$, where the mean $\boldsymbol \mu_k$, $\tau$ and the graph Laplacian $\boldsymbol L_k$ drive the heat diffusion processes. The number of signals is fixed to $M = 200$ and each experiment has been repeated 100 times.
		
\begin{figure}
			\centering
			\begin{subfigure}{.5\linewidth}
				\centering
				\includegraphics[width=\linewidth]{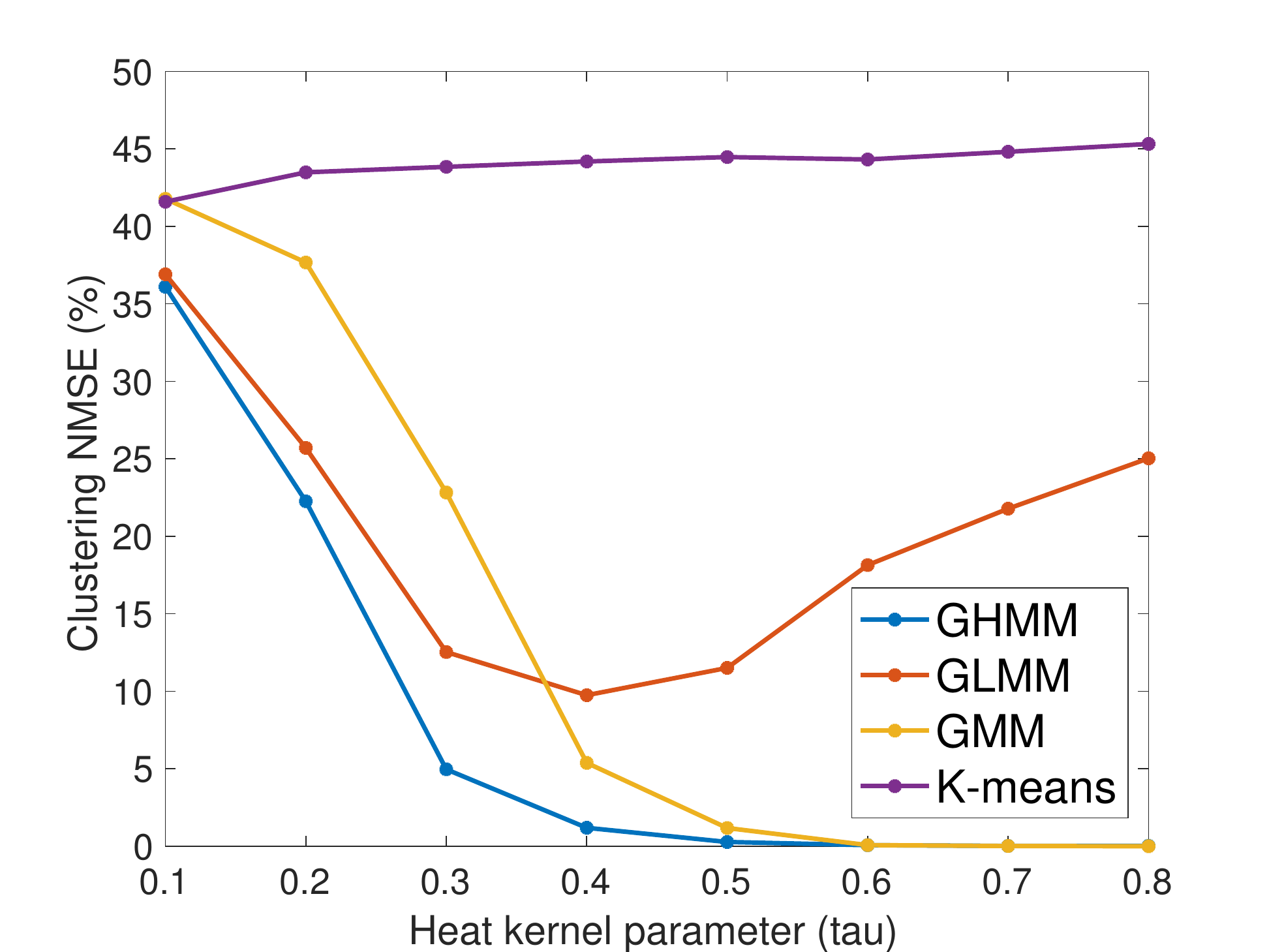}
				\caption{Clustering performance}
				\label{fig:tau_c}
			\end{subfigure}%
			\begin{subfigure}{.5\linewidth}
				\centering
				\includegraphics[width=\linewidth]{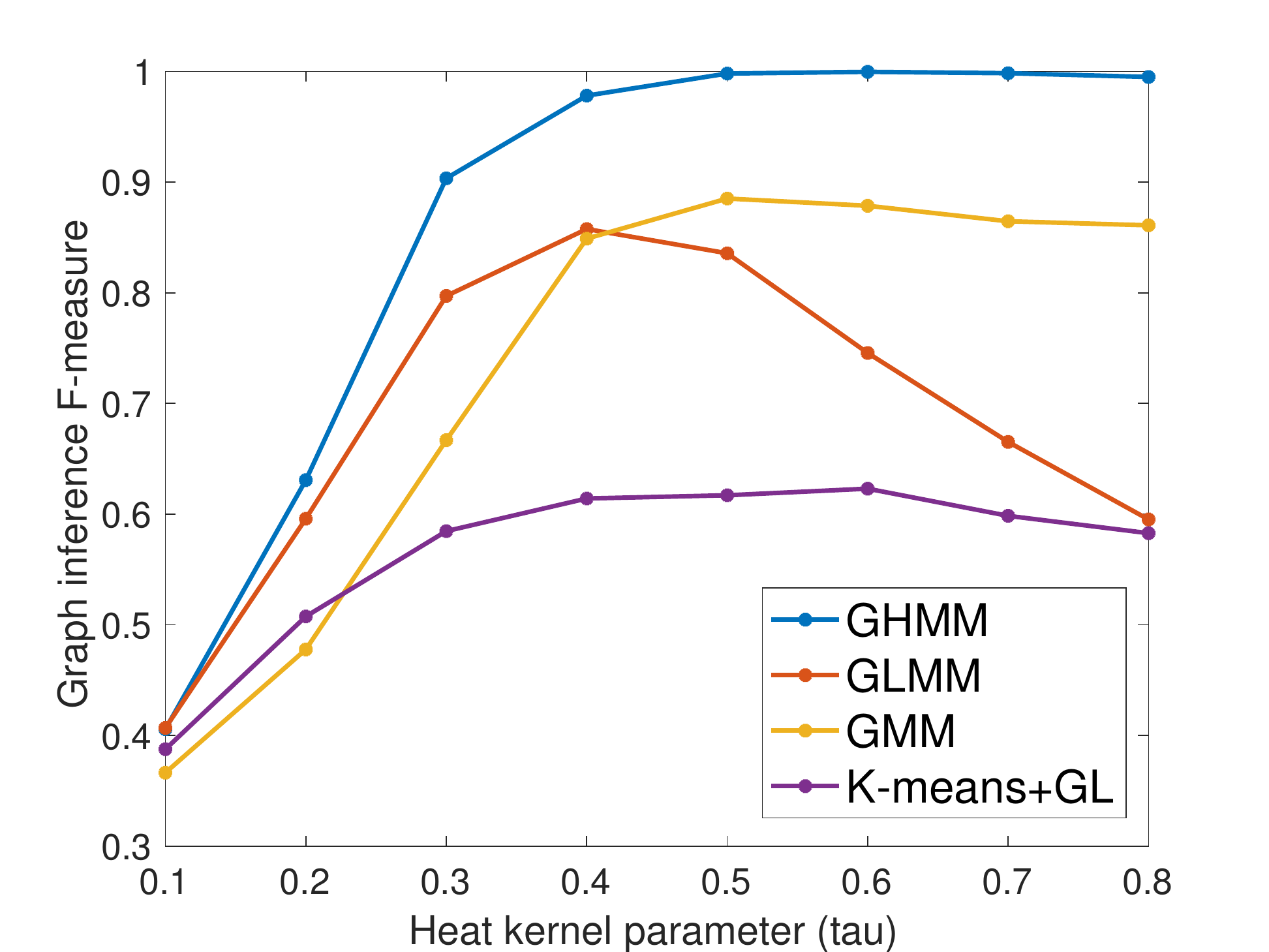}
				\caption{Graph inference}
				\label{fig:tau_f}
			\end{subfigure}
			\caption{Performance with respect to the heat parameter $\tau$.}
			\label{fig:tau}
		\end{figure}

		We test the performance of all four inference algorithms as a function of the heat parameter $\tau$ that varies between 0.1 and 0.8, as shown in Figure \ref{fig:tau}. For very small values of $\tau$, all algorithms have difficulties in recovering the structure as the sample covariance matrix is close to identity, and does not contain a lot of graph related information. For large values of $\tau$, the signals that we observe are very smooth, with very small variations. For this reason, the simple smoothness assumption used in GLMM is too weak to successfully separate signals, while the heat kernel model achieves very good performance. This shows that adapting the signal model to data can be very important, and emphasizes the importance of the flexibility our framework offers in incorporating various graph learning methods.
	
		Note that no prior information on $\tau$ was given to any of the algorithms (as it is just a scaling factor in the heat model). Therefore, the only influence that $\tau$ has on results comes from data structure.

		\subsubsection{Signals with noisy labels}
		Cluster labels are often not entirely unknown, but the available labels are unreliable due to some errors or noise. In this experiment, we examine this problem and evaluate the algorithm given noisy cluster labels. For the purpose of this experiment, we slightly modify our model, allowing for mixing coefficients to come in the form of group priors $\boldsymbol \alpha$. Namely, each signal $\boldsymbol x_m$ belongs to a predefined group $S_i$, and $p(z_{m,k} = 1) = \alpha_{S_i, k}$, if $\boldsymbol x_m \in S_i$. In this experiment, each signal has a noisy cluster label available. However, using the labels directly with a hard assignment might not be optimal due to the noise. We therefore group all the signals with the same noisy label, and vary the group prior initialisation: from 1 corresponding to the hard assignment of noisy labels, to 0.33 corresponding to a random prior . We use the same settings as in the experiment A.1) with 3 balanced clusters.
		
			\begin{figure}
			\centering
			\begin{subfigure}{.5\linewidth}
				\centering
				\includegraphics[width=\linewidth]{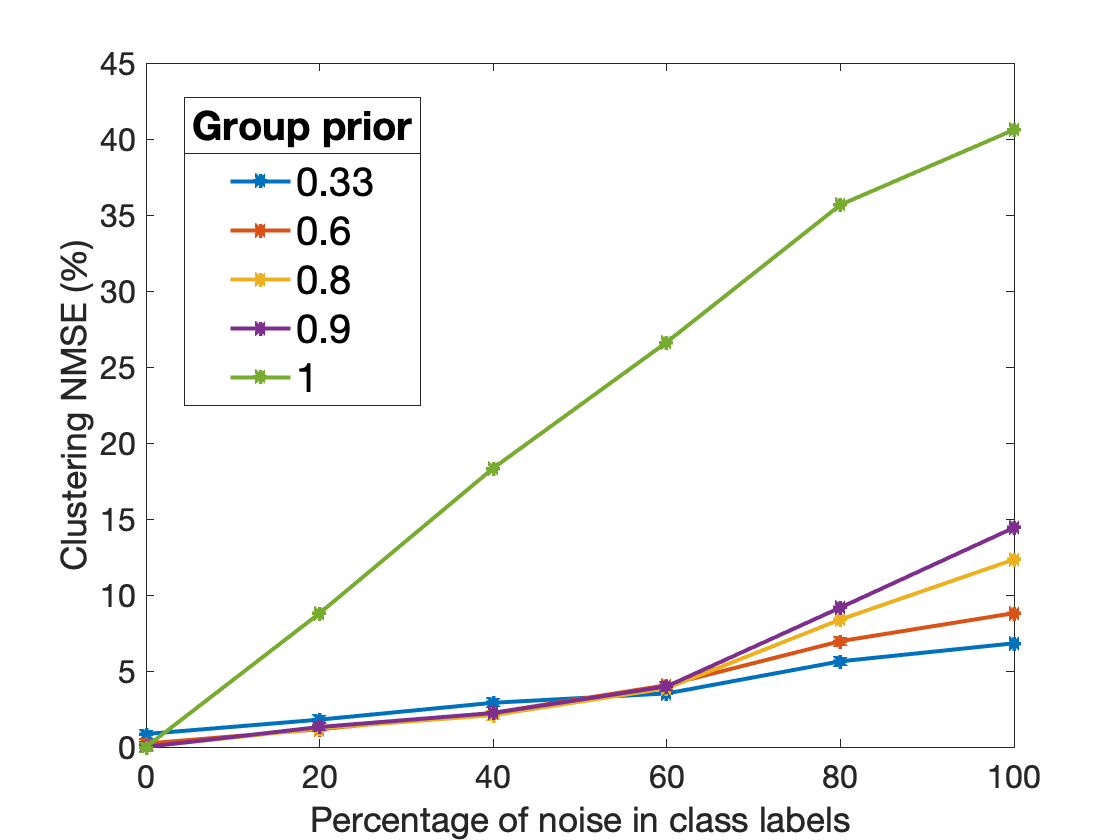}
				\label{fig:noisy_clustering}
			\end{subfigure}%
			\begin{subfigure}{.5\linewidth}
				\centering
				\includegraphics[width=\linewidth]{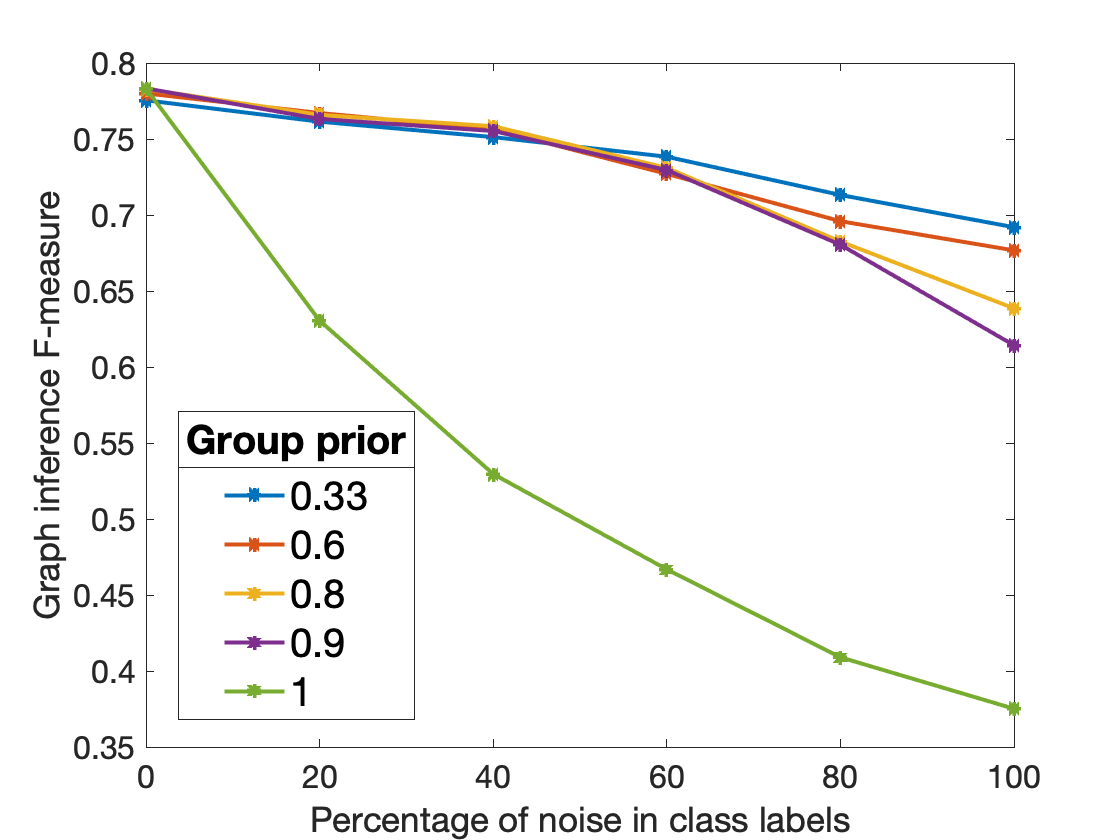}
				\label{fig:noisy_fmeasure}
			\end{subfigure}
			\caption{Performance with regard to the amount of noise present in data labels.}
			\label{fig:noisy_labels}
		\end{figure}
		
	Figure \ref{fig:noisy_labels} shows results in terms of clustering and graph inference F-measure for different percentages of noise among available labels (from 0 to 100). Note that, apart from the scenario in which the prior is fixed as 1 (hard assignment of labels), all other priors manage to adapt reasonably well, with 0.8 and 0.9 performing optimally for small noise levels, and 0.33 when noise is larger than 50\%. In addition, even small group priors benefit from scenarios with no noise, probably due to the fact that mere separation of signals into (almost) correct groups is enough to help the algorithm eventually converge to better clustering.
		\subsubsection{Gaussian mixture model signals}

\begin{figure}
			\centering
			\begin{subfigure}{.5\linewidth}
				\centering
				\includegraphics[width=\linewidth]{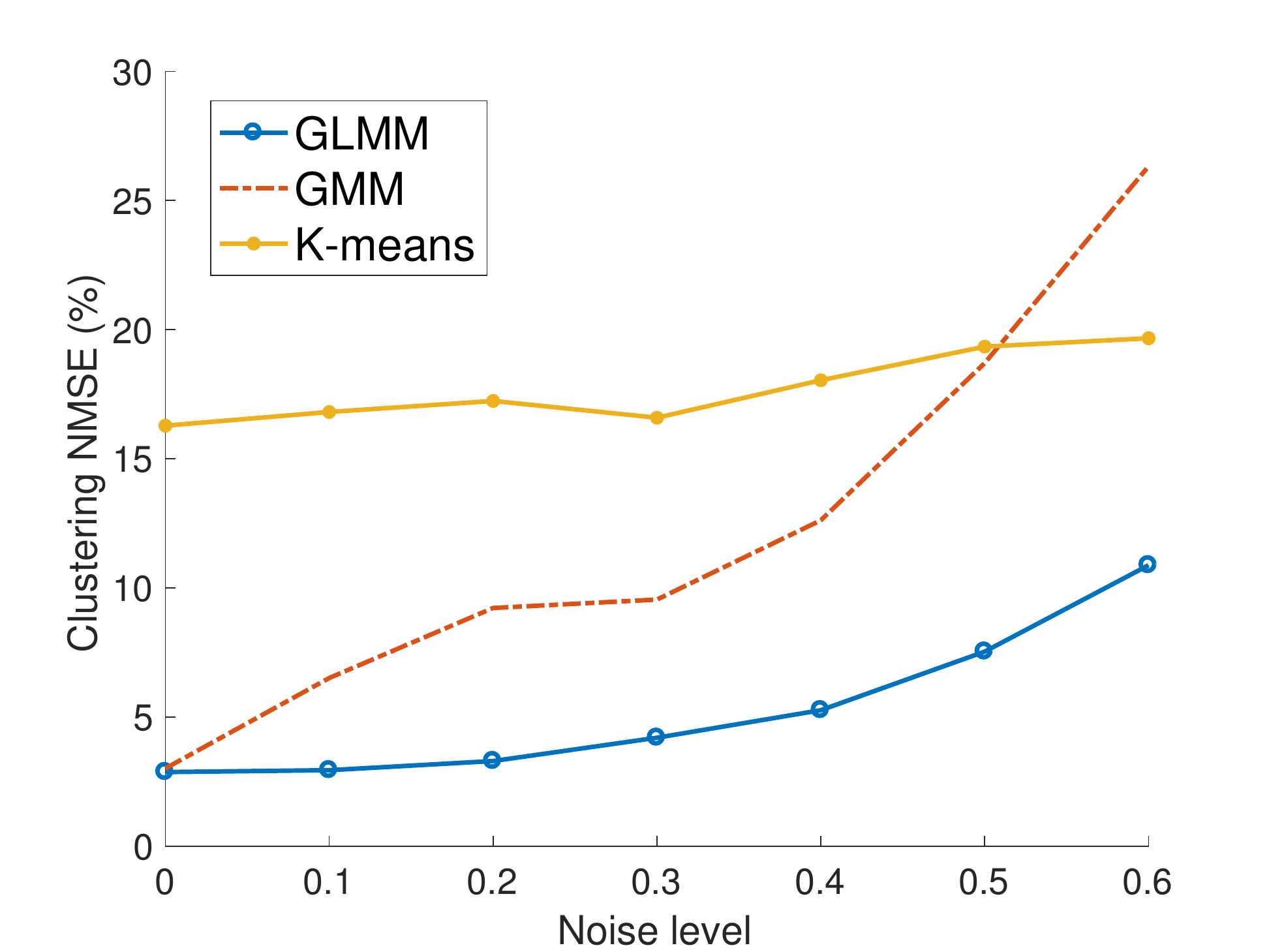}
				\label{fig:gmm_noise}
			\end{subfigure}%
			\begin{subfigure}{.5\linewidth}
				\centering
				\includegraphics[width=\linewidth]{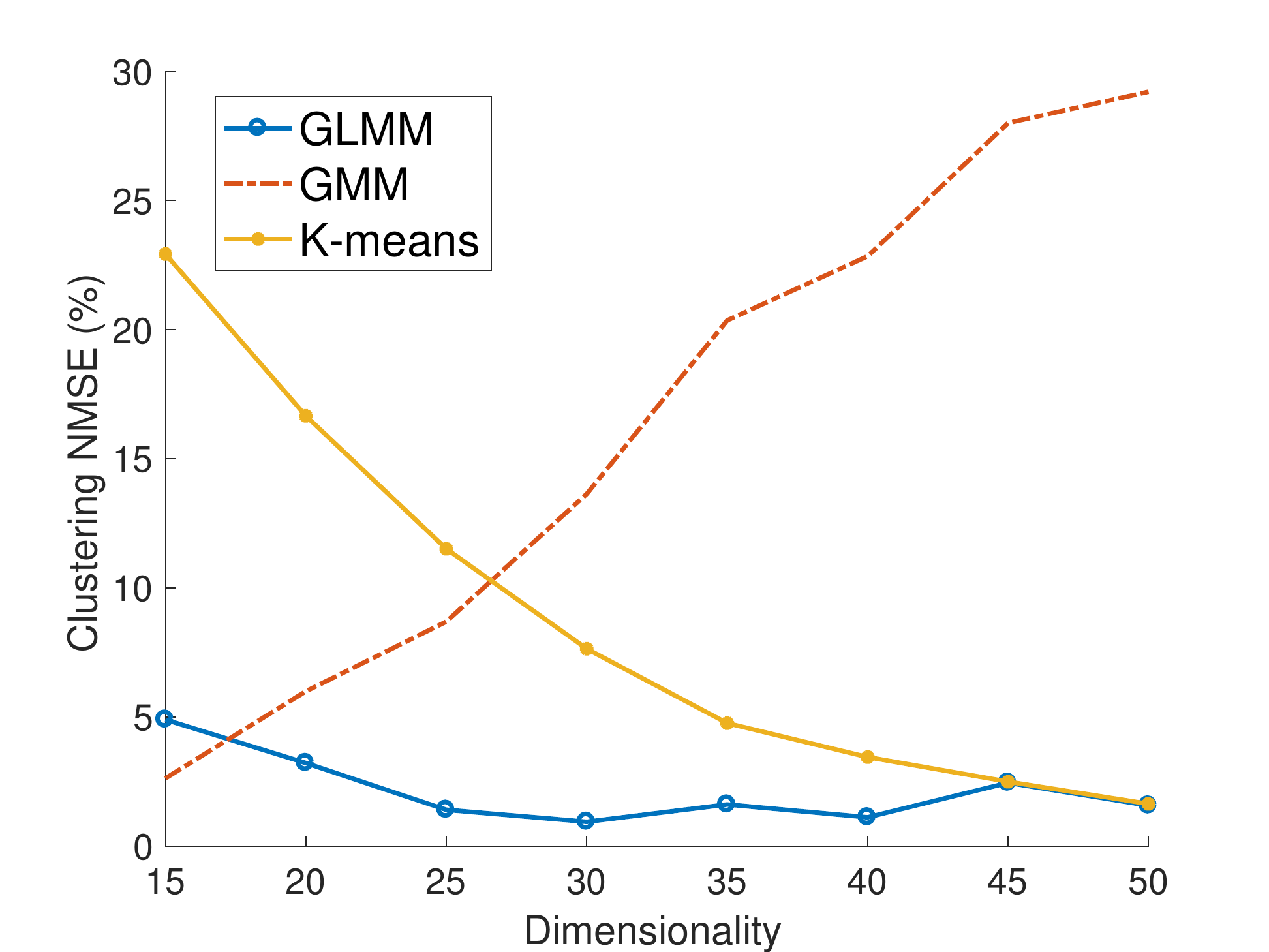}
				\label{fig:gmm_dims}
			\end{subfigure}
			\caption{Clustering performance of data generated through random Gaussian mixture models.}
			\label{fig:gmm}
		\end{figure}

		Finally, we test the performance of our algorithm on data generated from random Gaussian mixture models. The objective is to investigate if the added structure in GLMM creates implicit dimensionality reduction that can help in inference of high dimensional GMMs. 
		We generate random covariance matrices of size $N = 20$ from the Wishart distribution $\boldsymbol \Sigma_k 
		\sim \frac{1}{n} W (\mathbb{I}_n, n)$, and means as $\boldsymbol \mu_k 
		\sim \frac{1}{2} N (0, \mathbb{I}_n), \forall k$. With $\boldsymbol \alpha = [0.5, 0.5]$, we sample the Gaussian mixture model 200 times. We then add white noise to each sampled signal $\boldsymbol w \sim \mathcal{N}(0, \sigma^2 \mathbb{I})$, with $\sigma$ varying from 0 to 0.6, as in Figure \ref{fig:gmm}. Each point is averaged over 100 experiments. As expected, all methods are affected by increasing noise. However, it is interesting to see that, while the error of GMM and GLMM is very similar in the no-noise scenario, the GMM error increases drastically already for small noise ($\sigma = 0.1$). This is probably due to the high sensitivity to noise that Gaussian mixture models exhibit in high dimensions, while GLMM's implicit dimensionality reduction makes it more robust, even in cases when data is not sampled from a graph mixture. 
		
		To test this assumption further, we explore changing the dimensionality of the Gaussian mixtures. We vary $N$ from 15 to 50 as shown in Figure \ref{fig:gmm}, keeping the number of signals fixed (M = 200). Note that with larger $N$ the task becomes easier, as means are still generated in the same way ($\boldsymbol \mu_k \sim \frac{1}{2} \mathcal{N}(0, \mathbb{I}_n)$), so in higher dimensions cluster means are more likely to be further away. That explains a significant decrease in error for K-means. However, GMM still suffers from the curse of dimensionality, while GLMM manages to achieve lower errors for higher dimensions, implying an implicit dimensionality reduction has occurred.
		
		Various experiments on synthetic data show that our method achieves promising results in comparison with other inspected methods, introducing higher flexibility in data modelling and reducing the dimensionality of the problem.

		\subsection{Real data}
		We further evaluate our algorithm on real data. In applications where the real network is not known, we evaluate the performance by inspecting the clustering results. We further compare the inferred graphs to a constructed ground-truth graph, and use visual inspection to assess graph quality. 
		
		\subsubsection{Molene weather dataset}
		The Molene weather dataset \cite{girault2015signal} provides measurements of temperature and wind strength in 28 stations across Bretagne, France. There are in total 744 measurements of each type, in each of the 28 stations. Note that we refer to signals as measurements, while measure represents a whole class of temperature or wind strength signals.  We preprocess the data by subtracting the mean of each signal and by normalising both measures, to ensure the data is in the same range. We compare the results of GLMM, GMM, K-means + GL and GHMM in terms of clustering accuracy, graph inference, as well as model generalisability.
		
		We first look at clustering accuracy and graph inference performance. We randomly select a subset of 300 preprocessed measurements describing temperature and 300 describing wind speed to create a dataset for evaluation. The whole experiment has been repeated 100 times, each time with a different randomly selected subset of measurements. Clustering performance is presented in Table \ref{tb:molene} in terms of NMSE ($\%$).
		\begin{table} [h] 
			\caption{Clustering of 600 randomly selected signals from Molene weather dataset, of which 300 are temperature measurements and 300 represent wind speed. C stands for the clustering error in terms of NMSE (\%), while G stands for graph inference error in terms of edge recovery F measure.}
			\centering
			\begin{tabular}{|c||c|c|c|c|}
				\hline 
				& GLMM & GMM & K-means + GL & GHMM\\ 
				\hhline{|=||=|=|=|=|}
				C& \textbf{7.66} & 24.51 & 21.61 & 18.68 \\ 
				\hline
				G& \textbf{0.82} & 0.49 & 0.73 & 0.52\\
				\hline 
			\end{tabular} \label{tb:molene}
		\end{table}
		
		We can see that GLMM outperforms other methods in terms of clustering accuracy. This is especially interesting as the examined dataset does not have an inherent ground truth graph structure supporting it. We argue that this suggests that our data can be well modelled with graphs. Therefore, additional constraints posed by Laplacian inference reduce the scope of the problem when compared to GMM, while they still allow for a lot more adaptivity when compared to simple K-means. The results obtained by GHMM suggest that, while there is still added benefit compared to the other two methods, the choice of kernel in our framework is very important. Namely, the smooth kernel is more general and therefore more flexible to possible outside influences, while the heat kernel is a more strict model, and as such should be used more carefully with noisy data.
		
		We also investigate graph inference performance on this data. As there are no ground-truth graphs for this data, it is difficult to compare inference performance in a fair way. We thus construct pseudo-ground-truth graphs using the knowledge of original clusters. For each cluster separately, we use all available data, preprocess them by subtracting the mean, and use a graph learning algorithm \cite{kalofolias2016learn} to infer the graph. We present the results in Table \ref{tb:molene}. 
		While this is clearly not a conclusive comparison due to the lack of real ground truth graphs, it is interesting to see that there is a significant gain of GLMM compared to K-means + GL, both of which use the same graph inference method.
		
\begin{figure}
			\centering
			\begin{subfigure}{.5\linewidth}
				\centering
				\includegraphics[width=\linewidth]{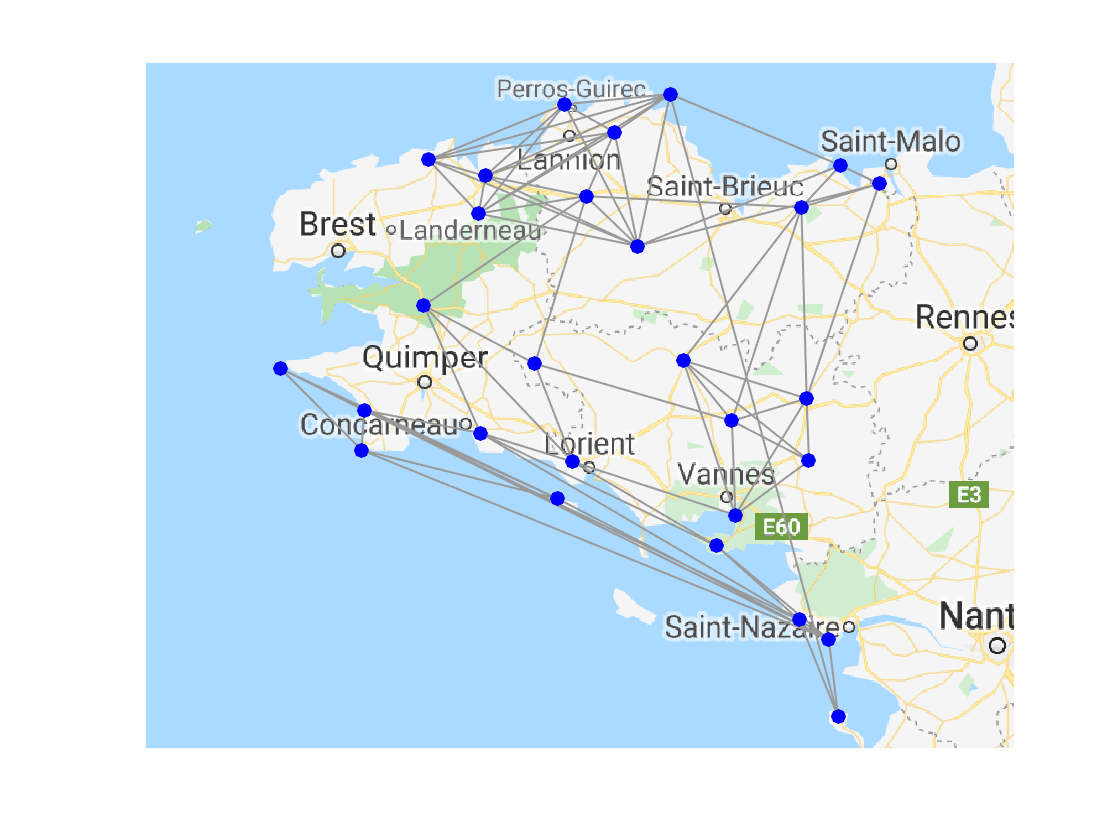}
			\end{subfigure}%
			\begin{subfigure}{.5\linewidth}
				\centering
				\includegraphics[width=\linewidth]{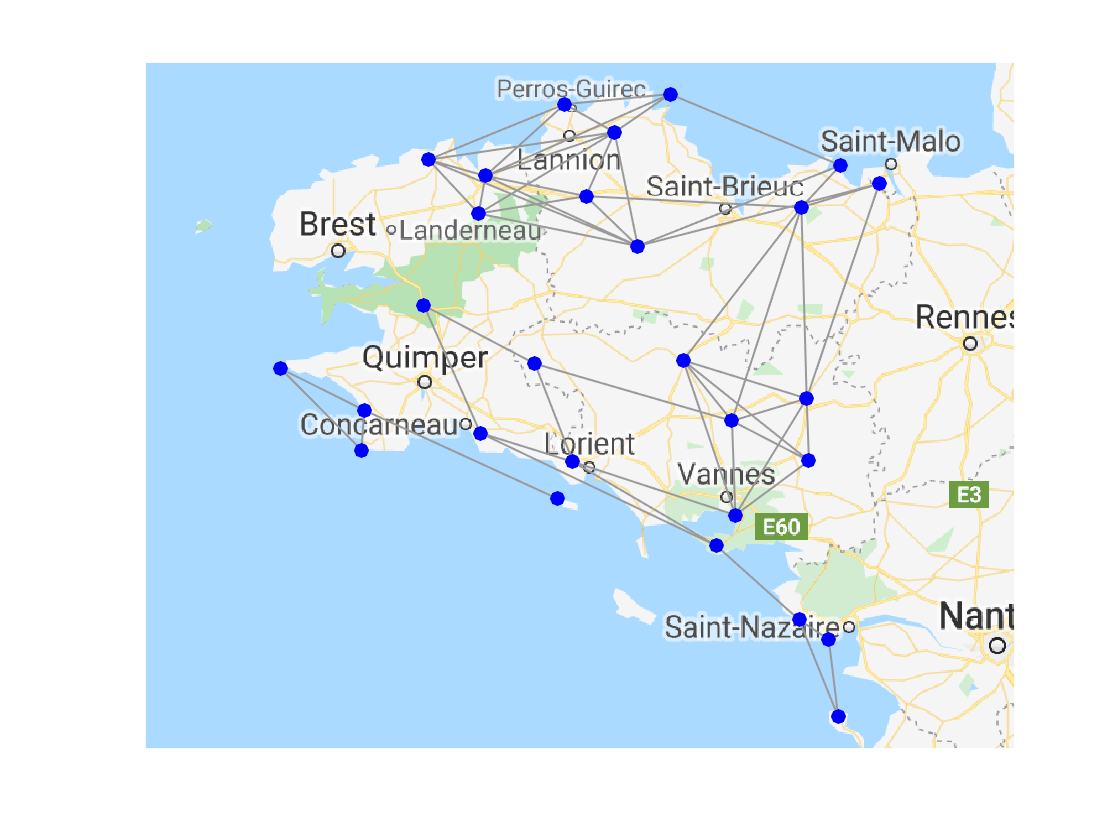}
			\end{subfigure}
			\caption{GLMM inferred temperature and wind graphs, respectively.}
			\label{fig:glmm_gr}
		\end{figure}

\begin{figure}
			\centering
			\begin{subfigure}{.5\linewidth}
				\centering
				\includegraphics[width=\linewidth]{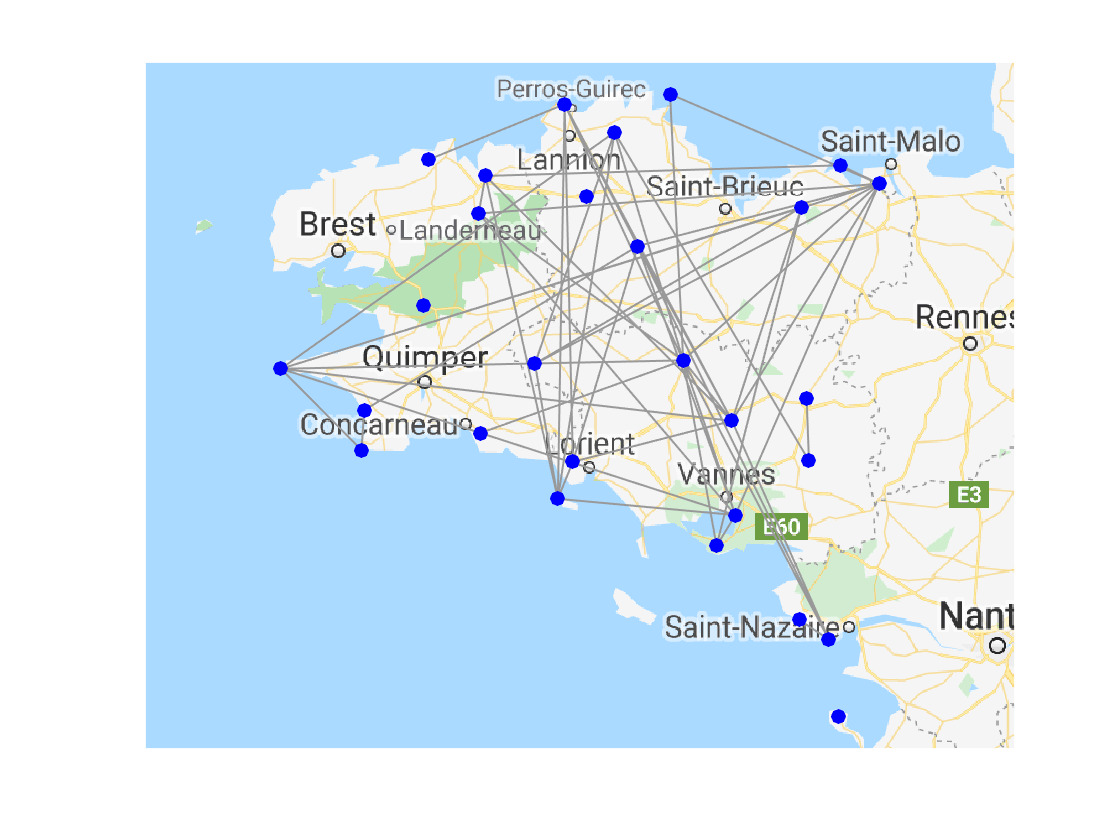}
			\end{subfigure}%
			\begin{subfigure}{.5\linewidth}
				\centering
				\includegraphics[width=\linewidth]{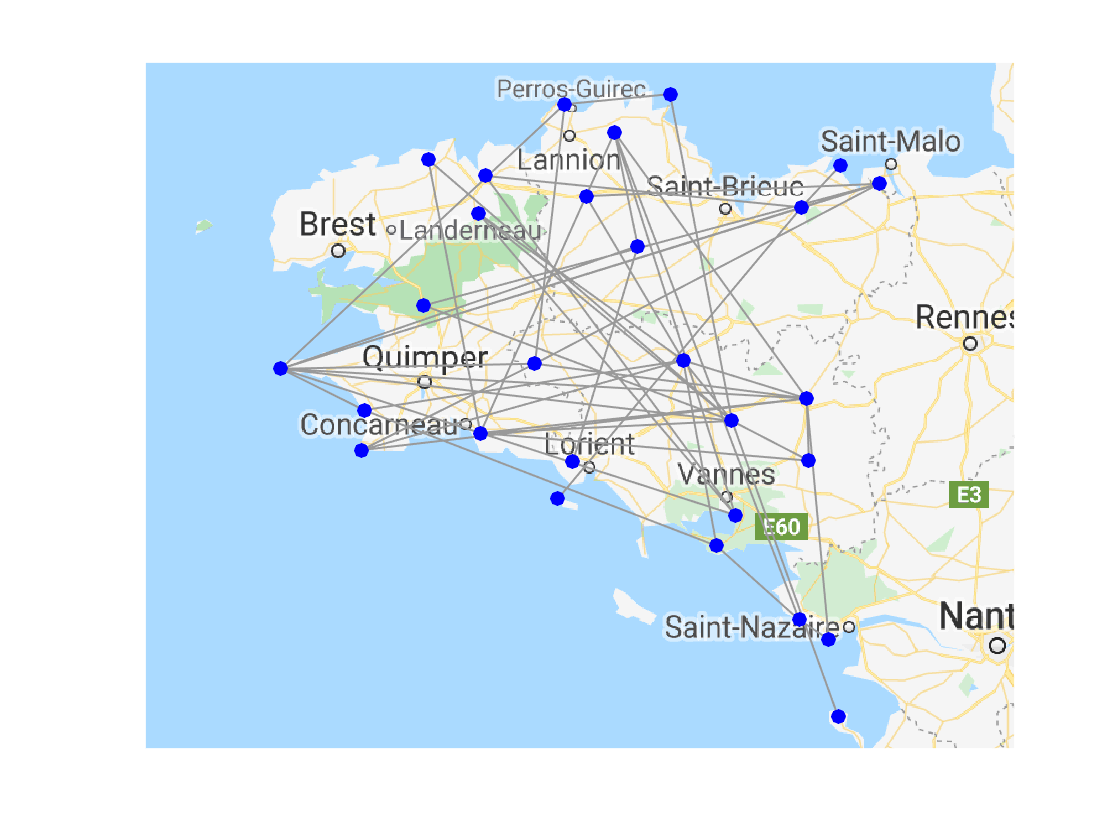}
			\end{subfigure}
			\caption{GMM inferred temperature and wind graphs, respectively.}
			\label{fig:gmm_gr}
		\end{figure}

		To complement numerical findings, we further investigate graph quality by visual inspection. Figure \ref{fig:glmm_gr} shows inferred temperature and wind graph topologies for GLMM. Figure \ref{fig:gmm_gr} presents the same for GMM. We note that these are geographical graphs, plotted with node position corresponding to true measuring station coordinates and that none of the algorithms were given any prior information on node positions. We can see that graphs inferred by GLMM offer much more structure, mostly connecting neighbouring nodes, a desirable property in a weather-related geographical graph. We also note that the temperature and wind graph inferred by GLMM are fairly similar, with the largest difference appearing along the southern coast. One possible explanation could be that the temperature values are all highly correlated as they are regulated by the ocean, while the wind values in these areas are less stable, often change direction and strength along the coast.

\begin{figure}
			\centering
			\includegraphics[width=0.7\linewidth]{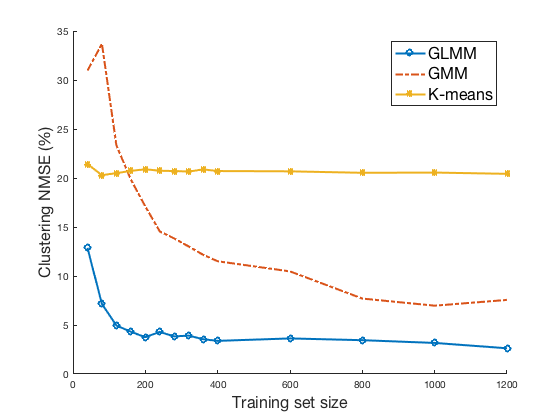}
			\caption{Test data clustering performance for different sizes of training data. Each training dataset contains 50$\%$ temperature and 50\% wind strength signals. The x-axis represents the training set size. The y-axis show signal clustering NMSE($\%$).}
			\label{fig:training}
		\end{figure}

		Next, we evaluate how trained models generalise to new data. Namely, we separate both temperature and wind data into 600 training and 100 testing signals. We then randomly choose subsets of training data of different sizes to fit generative models. We run each algorithm 5 times with different random initialisations, and select the best run in terms of training data clustering performance.
		The unseen (test) data is then clustered using trained models. Namely, we determine the cluster for each new datapoint by estimating which of the pre-trained clusters it fits into the best.
		The whole experiment has been repeated 100 times, each time with different randomly selected measurements. Figure \ref{fig:training} shows the test data clustering NMSE ($\%$) for all three methods, given different training set sizes. 
		
		We can see a significantly better performance in our method compared to the others in terms of generalisability to new data. We reiterate that this is especially significant as the results are demonstrated on data that does not inherently live on a graph. We argue that this shows the flexibility of our model, as it is able to well generalise to unseen data on signals that do not necessarily live on a mixture of graphs (but might be well representable with them). Furthermore, we note that, while increasing the size of our training set improves the performance of more adaptive algorithms, like GMM and GLMM, K-means does not show the possibility to adapt due to its very simple nature.

		\subsubsection{Uber data}
		We next search for patterns in Uber data representing hourly pickups in New York City during the working days of September 2014 \footnote{https://github.com/fivethirtyeight/ubertlc-foil-response}. We divide the city into 29 taxi zones, and treat each zone as a node in our graphs. The signal on these nodes corresponds to the number of Uber pickups in the corresponding zone. We fix the number of clusters to $K = 5$, following the reasoning in \cite{thanou2017learning}. 
		\begin{figure}[t]
			\centering
			\includegraphics[scale = 0.125]{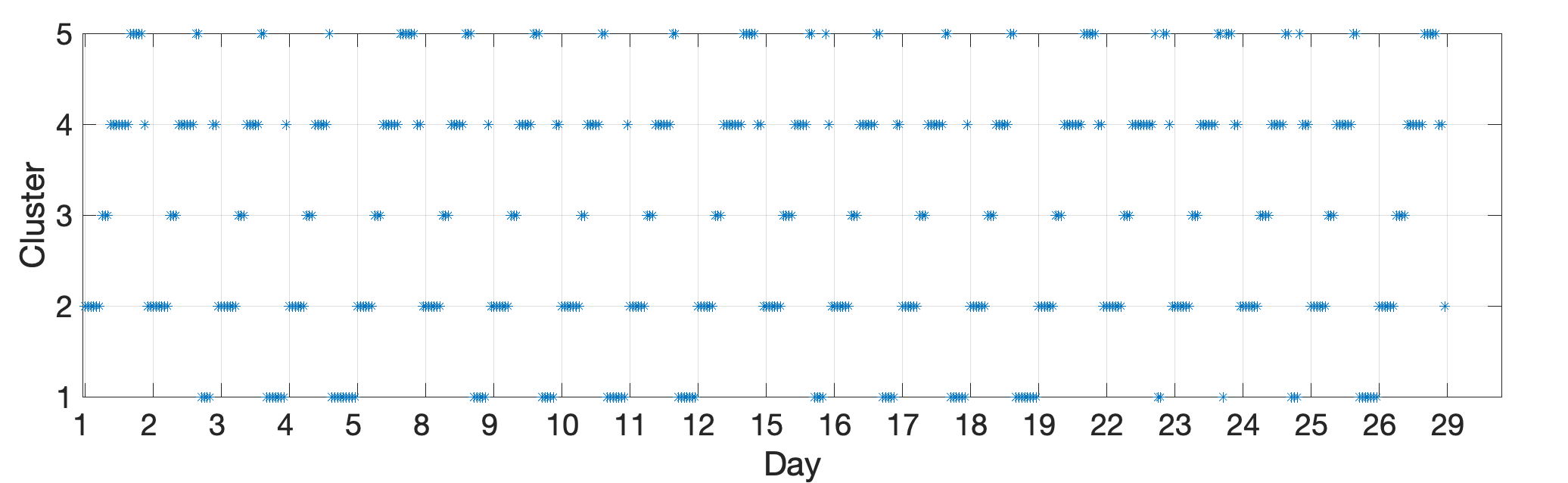}
			\caption{Cluster indexes for Uber hourly signals. Each dot represents one hour in the day, and thin vertical lines represent the beginning of each working day.}\label{fig:ghmm_uber}
		\end{figure}
		
		\begin{table} [h] 
			\caption{Clustering consistency error in terms of NMSE (\%).}
			\centering
			\begin{tabular}{|c|c|c|c|}
				\hline 
				GHMM & GMM & GLMM & K-means + GL\\ 
				\hhline{|=|=|=|=|}
				11.45 &    10.75
 & \textbf{10.07} & 13.55 \\ 
				\hline
			\end{tabular} \label{tb:uber}
		\end{table}
   
        Figure \ref{fig:ghmm_uber} shows the clustering of hourly Uber signals into 5 different clusters. We can see a slightly noisy periodic pattern, recurring daily. Note that no temporal information was given to the algorithm.
        
        As there are no ground truth clusters, we inspect the results in terms of clustering consistency. Namely, we compare clusters obtained in each day to clusters obtained in all other days and average the result, in order to approximate the expected difference between daily patterns. If the periodic pattern occurs in the same way every day, the clustering is consistent, and the average difference will be zero. Note that without ground truth clusters, consistency does not give sufficient information in evaluating clustering performance. Namely, a clustering that groups all data into only one cluster will be deemed perfectly consistent. For that reason, we complement our findings with qualitative analysis of clusters. Table \ref{tb:uber} shows the average NMSE obtained by comparing daily clusters. While GLMM has the best consistency, the result for GMM is not far in terms of consistency. However, GLMM on average separates the day into very sensible periods: 23h-6h, 6h-9h, 9h-15h + 22h-23h, 15h-20h and 20h-22h, separating night time, rush hours, day time and evening. At the same time, GMM on average separates the day into only 3 clusters, filling the remaining two clusters only with outliers, and putting most of the data into one large cluster: 6h-9h, 16h-20h and all the rest (20h-16h without 6h-9h). This explains a good score in consistency (as most data falls into the same cluster), however these clusters are not very meaningful. Note that GHMM and K-means produce clusters similar to GLMM, with the important difference of K-means not detecting morning rush hours (6h-9h and 9h-15h + 22h-23h are grouped into one cluster).

\begin{figure}[t]
			\centering
			\begin{subfigure}{.49\linewidth}
				\centering
				\includegraphics[width=\linewidth]{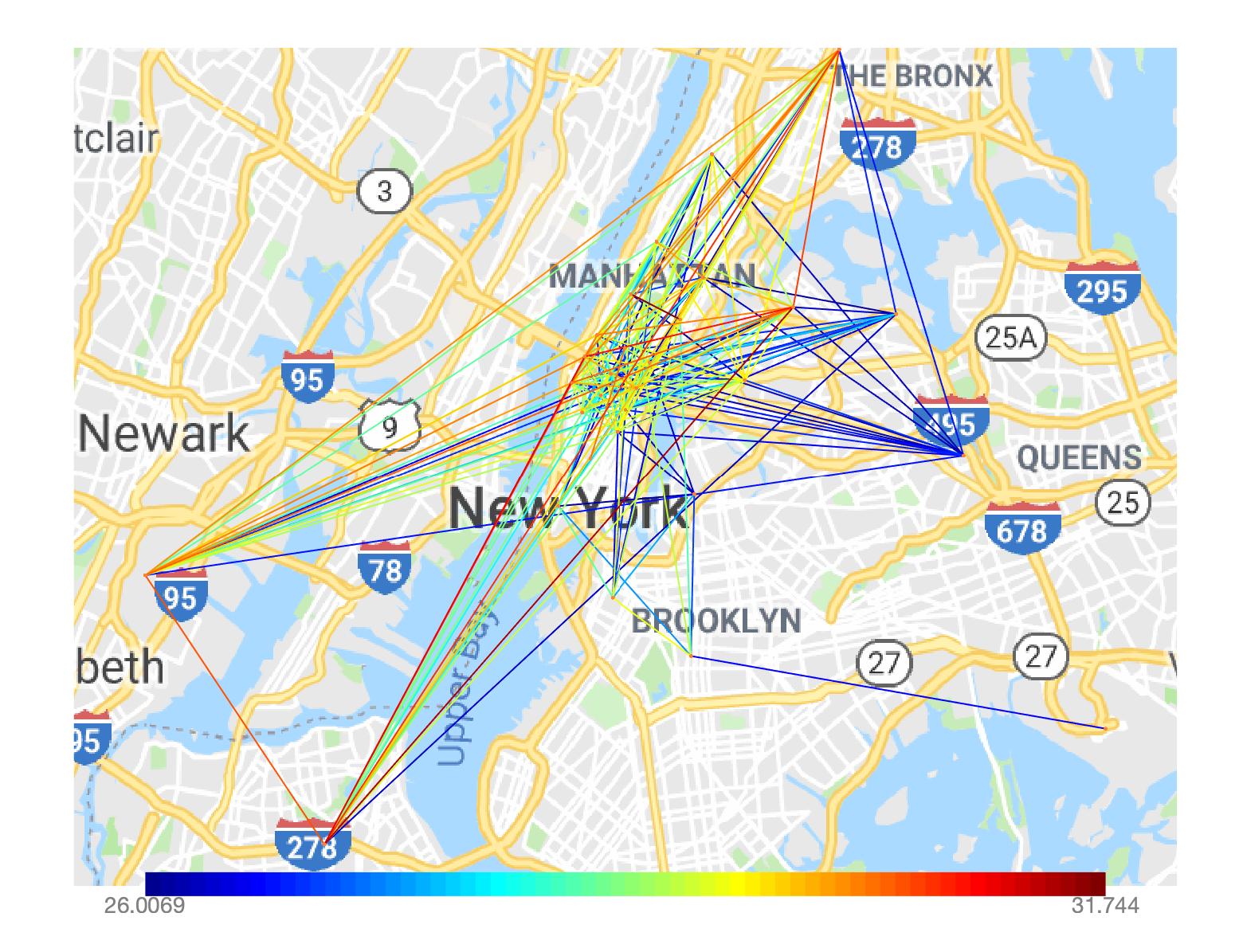}
				\caption{23h - 6h}
			\end{subfigure}%
			\begin{subfigure}{.49\linewidth}
				\centering
				\includegraphics[width=\linewidth]{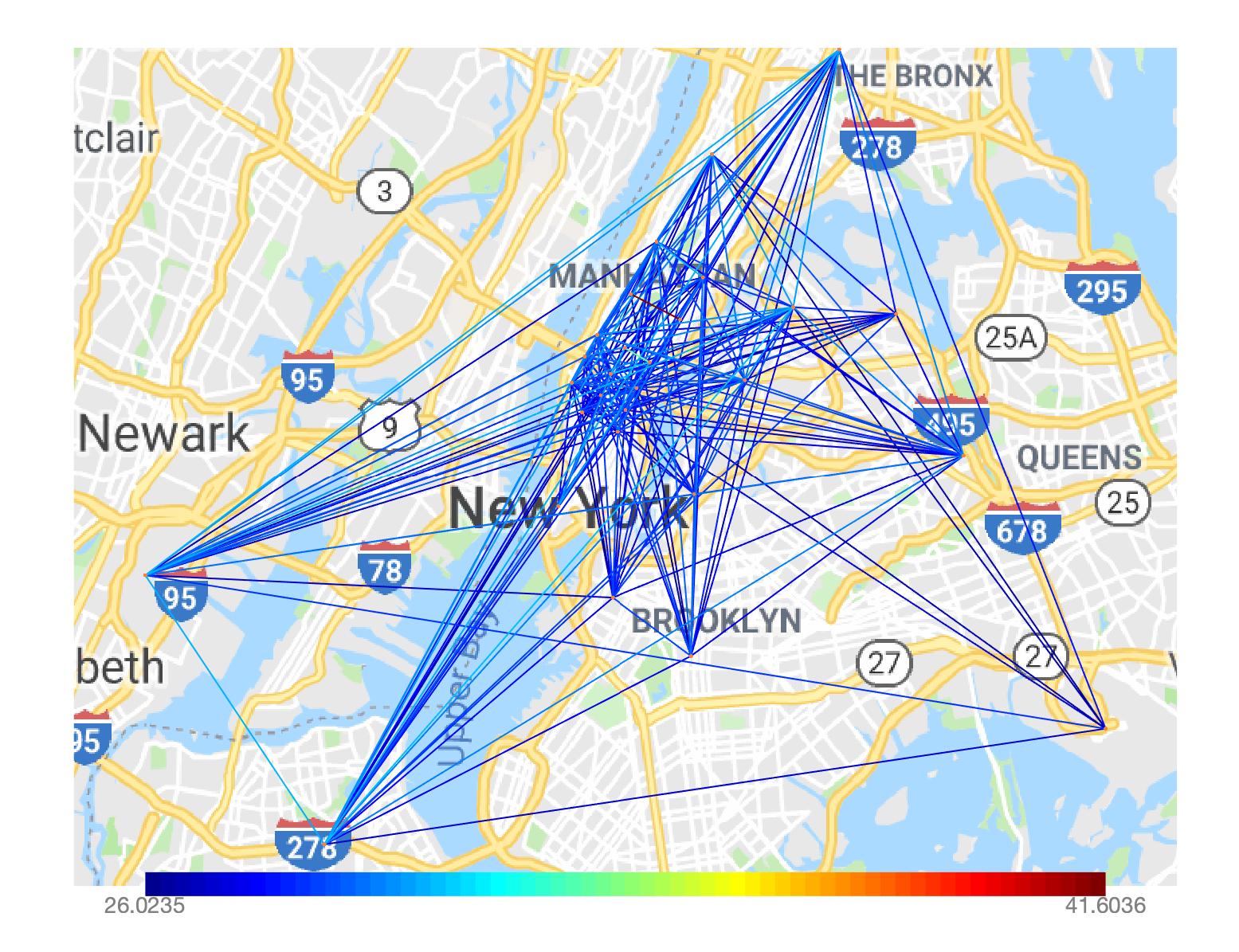}
				\caption{15h - 20h}
			\end{subfigure}
			\caption{Graphs corresponding to Uber patterns in different times of day.}
			\label{fig:uber_graphs}
		\end{figure}

		Finally, Figure \ref{fig:uber_graphs} presents 2 of the graphs inferred with GLMM. Each graph shows patterns of a different period in the day. We can see that the traffic during nights and early mornings is restricted to the city center and communications with the airports, while direct communication among non-central locations becomes more active later in the day. These different mobility patterns look reasonable with respect to daily people routine in NYC. 
	
	\subsubsection{MNIST digits}
		Finally, we comment on the advantage that our method brings over standard GMMs in terms of interpretability in high dimensional settings. We tackle a well known classification problem in which there is no reason to believe that signals live on a graph. Each signal is one MNIST digit representing "0" or "1". Since each digit is given as a 20x20 pixel grayscale image, the dimension of our signal is 400. We randomly choose 1000 digits from class '0' and 1000 digits from class '1'. We test the performance of GLMM, GMM and K-means + GL, as well as a modified GLMM that was restricted to allow edges only between 2-hop neighbouring pixels (mGLMM). We show clustering performance in Table \ref{tb:mnist}, but note that MNIST digit clustering is inherently a much lower dimensional problem, and should be treated as such. The experiments confirm this, giving better results for simpler methods: K-means performs very well as the simplest model, while GLMM still performs better than GMM due to its more focused nature and lower sensitivity to noise. The imposed structure in mGLMM simplifies the problem significantly and yields the best results, suggesting additional interpretative knowledge of the problem can be easily incorporated and reduce the dimensionality of the problem. Note that a mask of "allowed edges" was very easy to construct in this example, which is not necessarily always the case. However, these results imply that investigating contextual information might be a worthwhile task even when the masks are not as simple to construct. 
		\begin{table} [h] 
			\caption{Clustering of 2000 randomly selected MNIST digits, of which 1000 represent digit '0' and 1000 represent digit '1'. The error is presented in terms of NMSE (\%).}
			\centering
			\begin{tabular}{|c|c|c|c|c|}
				\hline 
				GLMM & mGLMM & GMM & K-means \\ 
				\hhline{|=|=|=|=|}
				1.76 & \textbf{0.64} & 7.18 & 0.89 \\ 
				\hline 
			\end{tabular} \label{tb:mnist}
		\end{table}

\begin{figure}
			\centering
			\includegraphics[width=0.6\linewidth]{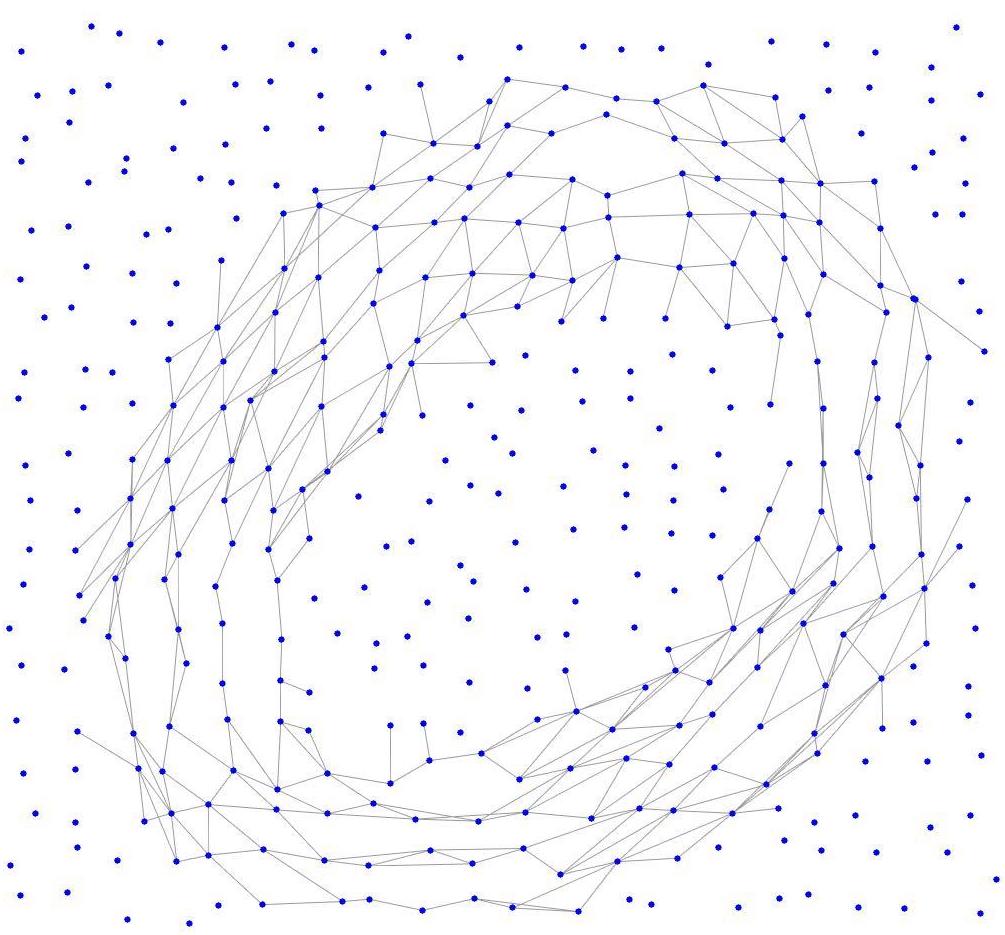}
			\caption{Recovered graph for the cluster of MNIST digit '0', no prior structure imposed}\label{fig:mnist_zero}
		\end{figure}

		At the same time, we discuss the interpretability advantage offered by GLMM in settings where this knowledge might not be available a priori, and can therefore not be imposed to the learning problem. Figure \ref{fig:mnist_zero} shows the graph topology inferred with GLMM corresponding to the cluster of digit zero. Note that no prior information or restrictions on edges were imposed in this version of the algorithm. To visualise the graph, we only considered edges adjacent to at least one vertex with a significant mean value larger than a threshold.
		As expected, we can see that neighbouring pixels that form the number zero are strongly correlated. We can also see that pixels are rarely connected if they are not close in the image, even though no pixel position information was given to the graph inference algorithm. It is clear that such insights can be very valuable in terms of interpretability in these very high dimensional settings.
		
		We finish with noting that the cost of using this model when data does not live on a graph comes, of course, in terms of restrictions imposed by a valid graph Laplacian, as well as the sparsity constraint in the graph learning method. These restrictions reduce model flexibility and could therefore lead to less accurate results. However, in large dimensional settings where this model is meant to be used, these restrictions are not too constraining. Even more, as they implicitly reduce the dimensionality of the problem, they seem to even ameliorate the performance on some datasets (as seen in Figure \ref{fig:gmm} and Table \ref{tb:molene}). 
		\section{CONCLUSION}
		We propose a generative model for mixtures of signals living on a collection of different graphs. We assume that both the mapping of signals to graphs, as well as the topologies of the graphs are unknown.
		We therefore jointly decouple the signals into clusters and infer multiple graph structures, one for each cluster.
		To do so, we formulate a Maximum a posteriori problem and solve it through Expectation minimisation. Our method offers high flexibility in terms of incorporation of different signal models, as well as simple inclusion of additional priors on the graph structure. We further argue that our model can be used for high dimensional clustering even when data is not assumed to have inherent graph structures, bringing additional interpretability to data. Simulations on real data achieve good clustering and meaningful graphs on weather data, infer interesting patterns in New York traffic, and find highly interpretable results on MNIST digits.

	\bibliographystyle{IEEEtran}
	\bibliography{references}

\end{document}